\documentclass[letterpaper, 10 pt, journal, twoside]{IEEEtran}
\IEEEoverridecommandlockouts                             
\usepackage[T1]{fontenc}

\usepackage[space, compress, sort]{cite}

\usepackage{amssymb} 
\usepackage{amsfonts}
\usepackage{amsthm}
\usepackage{pifont}
\usepackage{wrapfig}
\usepackage[ruled,vlined]{algorithm2e}
\usepackage{mathtools}
\usepackage{tabularx}
\usepackage{graphicx}
\usepackage{enumerate}
\usepackage{float}
\usepackage{url}
\usepackage{tikz}
\usepackage{verbatim}
\usepackage{booktabs} 
\usepackage{multirow}
\usepackage[linkcolor=black,citecolor=black,urlcolor=black,colorlinks=true]{hyperref}
\usepackage{bm}
\usepackage{subfigure}
\usepackage{algpseudocode}

\SetKwRepeat{Do}{do}{while}

\usepackage{xcolor}

\newcommand{\eg}{\textit{e.g.,}}
\newcommand{\ie}{\textit{i.e.,}}

\newcommand{\mc}[1]{{\mathcal{#1}}}

\usepackage{graphicx}
\usepackage[font=footnotesize]{caption}
\usepackage{cuted}
\usepackage{lipsum}

\let\oldtwocolumn\twocolumn
\renewcommand\twocolumn[1][]{%
  \oldtwocolumn[{#1
    \begin{center}
      \includegraphics[width=\textwidth]{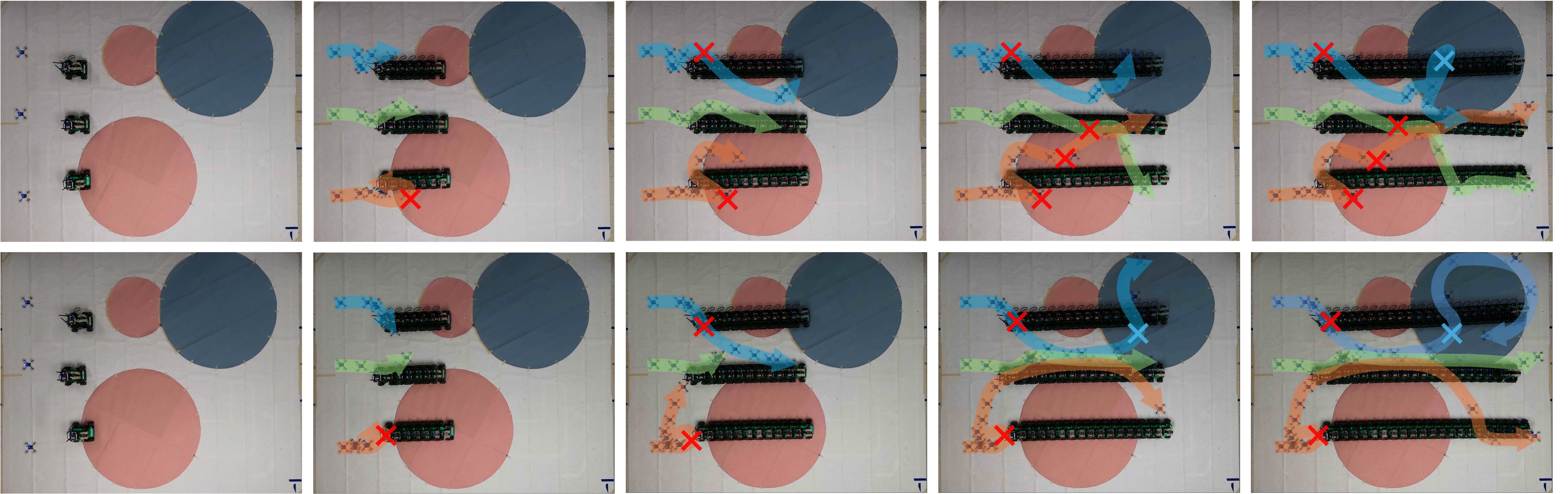}
      \captionof{figure}{
      Snapshots at different time steps showing multi-robot target tracking in environments with adversarial danger zones. 
      Three Crazyflie drones autonomously track three ground robots while navigating one communication danger zone (blue) and two sensing danger zones (red), which induce robot failures. 
      The top row illustrates scenarios with temporary failures, while the bottom row depicts permanent failures. 
      Arrows colored blue, green, and orange represent the trajectories of the individual drones. 
      Sensor attacks are indicated by {\color{red}{\ding{53}}} and communication attacks by {\color{cyan}{\ding{53}}}. }
      \label{fig:real_exp}
    \end{center}
    \vspace{0.1cm}
  }]
}

\title{Failure-Aware Multi-Robot Coordination for Resilient and Adaptive Target Tracking}

\author{Peihan Li$^{1}$, Jiazhen Liu$^{2}$, Yuwei Wu$^{3}$, Lifeng Zhou$^{1}$\textsuperscript{\textdagger} 
\thanks{$^{1}$Peihan Li and Lifeng Zhou are with the Department of Electrical and Computer Engineering, Drexel University, Philadelphia, PA 19104, USA. Email: \texttt{\small \{pl525,lz457\}@drexel.edu}.}
\thanks{$^{2}$Jiazhen Liu is with the Institute for Robotics and Intelligent Machines, Georgia Institute of Technology, Atlanta, GA 30332, USA. Email: \texttt{\small jliu3103@gatech.edu}.}
\thanks{$^{3}$Yuwei Wu is with the GRASP Lab, University of Pennsylvania, Philadelphia, PA 19104, USA. Email: \texttt{\small yuweiwu@seas.upenn.edu}.}
\thanks{\textsuperscript{\textdagger} Corresponding author.}
}

\begin{document}

\maketitle
\thispagestyle{empty}
\pagestyle{empty}

\begin{abstract}

Multi-robot coordination is crucial for autonomous systems, yet real-world deployments often encounter various failures.
These include both temporary and permanent disruptions in sensing and communication, which can significantly degrade system robustness and performance if not explicitly modeled. 
Despite its practical importance, failure-aware coordination remains underexplored in the literature.
To bridge the gap between idealized conditions and the complexities of real-world environments, we propose a unified failure-aware coordination framework designed to enable resilient and adaptive multi-robot target tracking under both temporary and permanent failure conditions.
Our approach systematically distinguishes between two classes of failures: (1) probabilistic and temporary disruptions, where robots recover from intermittent sensing or communication losses by dynamically adapting paths and avoiding inferred danger zones, and (2) permanent failures, where robots lose sensing or communication capabilities irreversibly, requiring sustained, decentralized behavioral adaptation. 
To handle these scenarios, the robot team is partitioned into subgroups. 
Robots that remain connected form a communication group and collaboratively plan using partially centralized nonlinear optimization. 
Robots experiencing permanent disconnection or failure continue to operate independently through decentralized or individual optimization, allowing them to contribute to the task within their local context.
We extensively evaluate our method across a range of benchmark variations and conduct a comprehensive assessment under diverse real-world failure scenarios. Results show that our framework consistently achieves robust performance in realistic environments with unknown danger zones, offering a practical and generalizable solution for the multi-robot systems community.
\end{abstract}

\section{Introduction}

Failures frequently emerge during real-world deployment, even in well-designed systems, complicating coordination and autonomy. 
Robots operating in dynamic and unstructured environments are especially vulnerable to disruptions and risks such as sensor degradation, communication loss, actuator faults, and adversarial interference~\cite{1435486, bezzo2014attack, 9430762, PORTUGAL20131572, 10.1145/3243734.3243752,  zhou2021multi, majumdar2020should}. 
These failures may occur temporarily, such as transient signal dropouts or occluded sensing, or permanently, due to hardware breakdowns or sustained environmental hazards. 
Despite increasing deployment of autonomous systems in high-stakes domains, most multi-robot frameworks assume fully functional sensors and reliable communication, making them fragile in failure-prone and adversarial conditions.

Several recent works have started to address risk or failure resilience in multi-robot coordination\cite{liu2022decentralized, mayya2022adaptive, 9349130, liu2024multi}. 
Existing works have approached this challenge through a range of strategies, including robustness to uncertainty~\cite{liu2024multi}, fault-tolerant control, redundant and continual task refinement~\cite{9812273}, and recovery mechanisms designed to mitigate partial system failures~\cite{Liresilient2024}. 
However, these methods typically address either sensing or communication disruptions in isolation, often assume prior knowledge of failure zones, and primarily focus on temporary failures.
Furthermore, few existing frameworks support flexible coordination modes that adapt in real time to evolving failures. 
This motivates the need for a more general, failure-aware coordination approach that explicitly models both temporary and permanent failures, and dynamically reconfigures multi-robot behavior under uncertainty.

To address these challenges, we propose a unified, resilient, and adaptive framework for multi-robot target tracking that is explicitly designed to operate under both temporary and permanent failures in sensing and communication. 
Our approach enables robots to dynamically switch between centralized coordination and decentralized decision-making based on network connectivity, while incorporating real-time inference of unknown danger zones that may trigger system-level disruptions. 
We validate our framework through extensive simulations and real-world experiments, demonstrating robust performance across a wide range of failure scenarios and highlighting its practical applicability in uncertain and adversarial environments.
To summarize, our contributions are as follows,

\begin{itemize}
    \item We propose a general resilient and adaptive multi-robot target tracking framework that formulates a partially centralized nonlinear optimization problem. The framework enables coordinated planning among communication-connected robots while allowing isolated robots to make decisions that contribute to overall team performance.
    
    \item The framework explicitly models both \textit{temporary and permanent failures in sensing and communication}, including those caused by adversarial attacks. Temporary failures capture intermittent disruptions, such as signal loss or sensor noise, while permanent failures represent irreversible compromises, including hardware damage or communication denial due to attacks.
    
    \item We extensively evaluate the proposed framework through a combination of simulations and real-world hardware experiments using aerial robots to track multiple ground vehicles. These evaluations demonstrate the effectiveness of the approach in maintaining tracking performance and system resilience and adaptiveness under various levels of uncertain and adversarial environments.
\end{itemize}

In what follows, related work is reviewed in Section II. Section III presents a system-level overview and discusses its generality across diverse application scenarios. Section IV provides a detailed description of the proposed approach, and Section V presents the performance evaluation and experimental results. We conclude the paper in Section VI.

\section{Related Works}

\subsection{Multi-Robot Target Tracking}
Multi-robot target tracking has emerged as a foundational problem in robotics, with applications ranging from environmental monitoring~\cite{dunbabin2012robots} and surveillance~\cite{rao1993fully} to search and rescue. 
A central challenge in this domain lies in coordinating multiple autonomous robots to maintain persistent and accurate tracking of dynamic targets, often under uncertainty and resource constraints. 
Previous work in this area focused on centralized tracking systems with known target dynamics, leveraging techniques such as Kalman filtering and model predictive control~\cite{robin2016multi,hausman2015cooperative}. 
However, recent trends have shifted toward decentralized and distributed approaches that offer greater scalability, robustness, and adaptability in complex, real-world settings~\cite{liu2022decentralized,10644530}. These developments have been further driven by advances in multi-robot planning~\cite{10644567}, communication-aware coordination, and resilience to adversarial disruptions~\cite{zhou2023robust,10802509}. In particular,~\cite{10644530} addresses multi-robot target tracking in 3D and proposes a method that integrates distributed target state estimation with the cooperative visual inertial odometry algorithm.
~\cite{10644567} introduces a meta-algorithm to learn whether it should follow external untrustworthy commands or a submodular coordination algorithm. The meta-algorithm outperforms either option alone and demonstrates how to improve robustness against untrustworthy commands.
Additionally,~\cite{10770001} investigates human-in-the-loop scenarios, comparing heuristic and reinforcement learning-based strategies for following a moving person. 
These works highlight a trend toward robust, scalable, and adaptive robot tracking in complex and uncertain environments.

\subsection{Temporary and Permanent Failures in Multi-Robot Systems}

In dynamic and partially known environments, multi-robot systems are often subject to temporary disruptions, such as intermittent sensor outages, communication delays, or abrupt target maneuvers, as well as permanent failures, including actuator faults, energy depletion, or the complete loss of robots mid-mission. 
While some failures may not immediately jeopardize the mission, they can propagate uncertainty, disrupt coordination, and reduce task effectiveness if not promptly mitigated.
~\cite{prorok2021beyond} provides a comprehensive taxonomy of the ways toward overcoming these stressors. Effective deployment in these settings requires planning and control strategies that can detect, interpret, and respond to disturbances with low latency and limited reliance on centralized infrastructure. 

Recent work has explored online planning strategies that incorporate dynamic obstacle avoidance, real-time kinematic feasibility, and robustness to sensing noise and delays~\cite{8460863, 9561759}. These frameworks typically emphasize local adaptation through feedback control, enabling robots to adjust their behavior based on evolving estimates of both environmental structure and team state. While effective for handling short-lived anomalies, many of these approaches presume persistent robot functionality and are not designed to absorb the effects of partial team failures or dynamic resource loss.

To address more severe disruptions, several methods integrate resilience into distributed planning and task reallocation. 
Robot dropout detection has been integrated with online task redistribution frameworks that utilize event-triggered replanning and game theory to dynamically reallocate coverage tasks in response to unexpected robot failures~\cite{Gupta2019}.
Utility-based goal reassignment has been applied in heterogeneous teams, using optimization-based task allocation under uncertainty to compensate for degraded robots~\cite{10802510}. Probabilistic models, particularly those using belief-space planning or chance-constrained optimization, anticipate risk and guide robot behavior under bounded uncertainty, enabling robust performance despite incomplete information~\cite{10333309}.

\subsection{Resilience and Adaptiveness in multi-robot systems}
Achieving robust performance beyond controlled environments requires resilience and adaptiveness. Towards this end, algorithmic designs for multi-robot systems generally fall into two categories. The first focuses on coordination strategies that \textit{prepare} the system to withstand attacks and failures~\cite{8534468,7962962}. The second category involves adaptive and reactive algorithms that allow systems to \textit{recover} from faults or adversarial events, as exemplified by~\cite{8968611,9196961}.

Specifically for multi-robot target tracking, maintaining long-term tracking performance under adversarial conditions poses challenges due to sensor faults and interrupted communications. 
Earlier works have explored embedding resilience directly into trajectory optimization or decision-making processes. 
For instance, Mayya et al.~\cite{mayya2022adaptive} proposed a risk-aware planning strategy that explicitly balances tracking performance and minimizing failure risk. 
Liu et al.~\cite{liu2022decentralized} utilized the consensus mechanism to decentralize this model, thereby improving its scalability. 
Zhou et al.~\cite{zhou2023robust} achieved resilience through a worst-case robust control formulation, providing theoretical guarantees even in the face of an arbitrary and unbounded number of attacks on sensing or communication links. Other studies~\cite{okumura2023fault, 10008951, 9197243} implemented fault-tolerant motion planning using control barrier functions, local fault diagnosis, or graph-theoretic redundancies, ensuring robustness to isolated failures in networked systems. These approaches emphasize \textit{preemptive mitigation} based on known or predictable risk, but often struggle to adapt to dynamic or unanticipated changes.

A parallel line of work focuses on \textit{reactive} strategies that facilitate recovery after attacks or failures occur. Such methods aim to maintain situational awareness and control effectiveness through online adaptation. For instance,~\cite{prorok2021beyond} discusses resilience through redundancy and mobility in decentralized information fusion, where robots can take over tasks from failed peers without explicit replanning. Similarly, adaptive control methods have been used to maintain formation integrity and communication connectivity in adversarial settings~\cite{8593630, 8534468, cavorsi2022multi, 9354993}. These methods often utilize online learning and parameter estimation to dynamically adjust behavior.

In scenarios where environment maps or hazard sources are initially unknown, Schwager et al.~\cite{schwager2017multi} proposed a mutual information-based exploration and tracking controller that can identify hazard locations while performing the primary task. Their strategy enables performance recovery through gradual exploration, allowing the robots to map out and circumvent threats as they are discovered. Recent advances have extended these ideas to hybrid systems that switch between safe and aggressive modes depending on environmental cues or predicted hazards~\cite{Abouelyazid_2023, 10234549}. However, many reactive strategies lack formal performance guarantees and become computationally expensive in large-scale deployments, especially under partial observability.

\begin{figure*}[ht]
    \centering
    \includegraphics[width=0.98\linewidth]{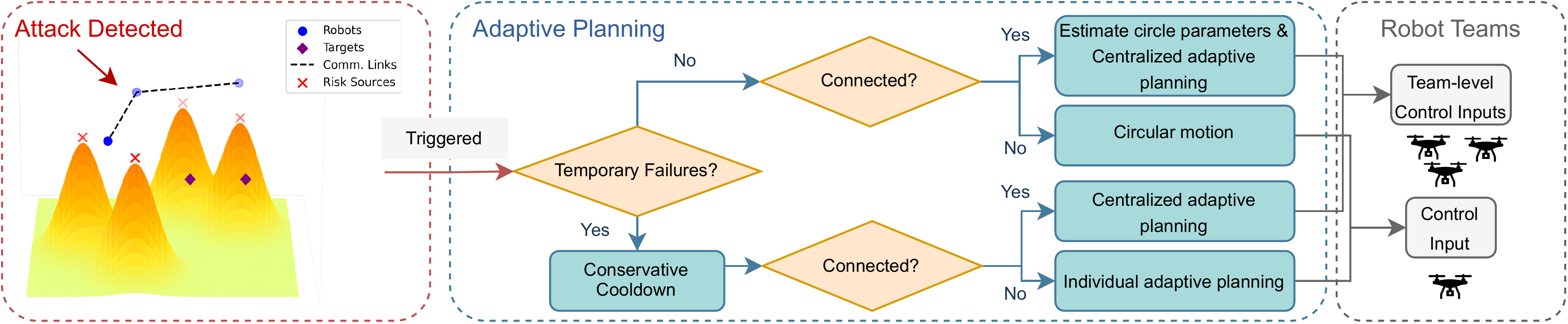}
    \caption{Flowchart for the coordination strategies under different attacks. We design adaptive planning strategies as specified in the middle part. The robots' adaptive behaviors are dependent on whether the failures are recoverable and whether the robot under attack maintains normal communication. The communication ability permits robots to plan as a team, while an isolated member performs individual planning. }
    \label{fig:framework}
\end{figure*}

\section{Backgrounds}

\subsection{Multi-Robot Target Tracking under Danger Zones}
\label{sec:background-basic}
The robot team’s mission is to collectively estimate and track the positions of dynamic targets using onboard range and bearing sensors and inter-robot communication. 
We consider a team of $M$ robots tracking a set of $N$ moving targets. Each robot is indexed by $i\in \mathcal{R}$, where $\mathcal{R}$ denotes the whole robot team. Robot $i$ maintains its state $\mathbf{x}_{i,t}$ and executes control input $\mathbf{u}_{i,t}$ at time step $t$. For each target $j$, where $j\in [1, 2, \cdots, N]$, we represent its (true) state at time $t$ as $\mathbf{y}_{j, t}$, and its estimated state as $\hat{\mathbf{y}}_{j, t}$. We describe the dynamics of targets collectively as,
\begin{equation}
    \mathbf{y}_{t} = \mathbf{A}\mathbf{y}_{t-1} + \mathbf{B} \mathbf{v}_{t-1} + \mathbf{w}_{t-1},
\end{equation}
where $\mathbf{A}$ is the state transition matrix, $\mathbf{B}$ is the control matrix, and $\mathbf{w}_{t-1}$ is the zero-mean Gaussian noise with covariance matrix $\mathbf{Q}$, \ie~$\mathbf{w} \sim \mathcal{N}(\mathbf{0}, \mathbf{Q})$. We denote the collective target state as $\mathbf{y}_{t} = [\mathbf{y}_{1, t}; \mathbf{y}_{2, t}; \dots; \mathbf{y}_{N, t}]$, with corresponding control inputs $\mathbf{v}_{t-1}$.

The robot team's observation of targets at time $t$ is calculated as
\begin{equation}
    \mathbf{z}_t = \mathbf{H}_t \mathbf{y}_t + \mathbf{\eta}_t,
\end{equation}
where $\mathbf{H}_t$ is the measurement matrix. Since the robots are equipped with inherently nonlinear range and bearing sensors, we linearized their measurement models to obtain $\mathbf{H}_t$. Please refer to~\cite{5735231} for details of the linearization. $\mathbf{\eta}_t$ is the measurement noise which follows a Gaussian distribution $\mathcal{N}(\mathbf{0}, \mathbf{R}_t)$, and we adopt the definition of $\mathbf{R}_t$ from~\cite{mayya2022adaptive} such that the covariances exponentially increase with the distance between the robot and the target. This reflects the realistic characteristics of measurements, as measuring from a longer distance often leads to higher inaccuracy. 

Multi-robot target tracking is formulated as an optimization problem that jointly minimizes tracking error and control effort.
Mathematically, we solve for robots' control input at every time step, balancing between the minimization of tracking error and control efforts, with the following program: 

\vspace{-0.2cm}
\begin{subequations}
\label{eq:opt_formulation}
\begin{align}
        \min_{\mathbf{u}_{i, t}, \forall i \in \mathcal{R}} \ \ &   w_1  \sum_{i=1}^{M} f(\mathbf{x}_{i, t+1}, \hat{\mathbf{y}}_{i,t+1}) + w_2 \sum_{i=1}^{M}\lVert \mathbf{u}_{i,t}\rVert \quad \ \ \ \label{eq:opt_objective}\\ 
        \textrm{s.t.} \
        & \mathbf{x}_{i, t+1} = \mathbf{\Phi}_i\mathbf{x}_{i,t} + \mathbf{\Lambda}_i\mathbf{u}_{i,t},  \forall i \in \mathcal{R}, \label{eq:opt_dynamics1} \\
        &  h(\mathbf{x}_{i,t+1}; \mathcal{E}_t) \leq 0, \quad \forall i \in \mathcal{R}, \label{eq:opt_env_func}
\end{align}
\end{subequations}
where \( f(\cdot) \) is a generic function that quantifies the tracking error of targets. The weights \( w_1 \) and \( w_2 \) determine the trade-off between tracking accuracy and total control effort. $\mathbf{x}_{i,t+1}$ is the position for robot $i$ at step $t+1$, which satisfies the dynamics constraint in Eq.~\ref{eq:opt_dynamics1} with state transition matrix $\mathbf{\Phi}_{i}$ and control matrix $\mathbf{\Lambda}_{i}$, respectively. 
Eq.~\ref{eq:opt_env_func} defines the environmental constraints through the function $h(\cdot; \mathcal{E}_t)$ that captures how the environmental condition $\mathcal{E}_t$ influences robots' behaviors. 
Such influence can be generic; for instance, it could encompass requirements such as collision avoidance or specific cautious strategies around hazardous sources. 
In this paper, we assume that robots (e.g., drones) operate in free-flying space without static obstacles, and we consider danger zones as environmental constraints of concern, which probabilistically induce sensing and communication failures that threaten the safety of the multi-robot team. Since robots operate at different altitudes, inter-robot collisions are not considered in this formulation, though they can be incorporated if needed.

We adopt the probabilistic definition of danger zones as introduced in~\cite{Liresilient2024, liu2024multi}. 
Let the environment contain \( p \) sensing danger zones, denoted by \( \{ \mathcal{S}_l \}_{l=1}^{p} \), and \( q \) communication danger zones, denoted by \( \{ \mathcal{C}_k \}_{k=1}^{q} \), both of which are \textit{initially unknown} to the robot team. 
The sensing danger zone $\mathcal{S}_l$ is centered at a random position $\mathbf{x}_{\mathcal{S}_l} \sim \mathcal{N}(\boldsymbol{\mu}_{\mathcal{S}_l}, \boldsymbol{\Sigma}_{\mathcal{S}_l})$ with safety radius $r_l$. The region where robot $i$ is at risk of sensing failure from zone $\mathcal{S}_l$ is defined as:
\begin{equation}
\mathcal{S}_l = \{ \mathbf{x}_i \in \mathbb{R}^{n_x} : \lVert \mathbf{x}_i - \mathbf{x}_{\mathcal{S}_l} \rVert \leq r_l \}.
\end{equation}
The communication danger zone $\mathcal{C}_k$ is defined by its center, whose position also follows a Gaussian distribution $\mathbf{x}_{\mathcal{C}_k} \sim \mathcal{N}(\boldsymbol{\mu}_{\mathcal{C}_k}, \boldsymbol{\Sigma}_{\mathcal{C}_k})$. We consider the communication channel between robot $i$ and robot $j$ to be jammed when the following condition holds,
\begin{equation}
\mathcal{C}_k = \{ \mathbf{x}_i \in \mathbb{R}^{n_x} : \lVert \mathbf{x}_i - \mathbf{x}_{\mathcal{C}_k} \rVert \leq \delta_2 \lVert \mathbf{x}_i - \mathbf{x}_j \rVert \},
\end{equation}
where $\delta_2 \in \mathbb{R}_{>0}$ is a tunable parameter related to signal strength and vulnerability.
If the environment were fully known in advance, with danger zone locations and characteristics specified a priori, they could be directly utilized for planning using an optimization program similar to~\cite{liu2024multi}, such that the robot team adopts a pre-cautious strategy to mitigate potential failures.  

\subsection{Failures in Unknown Danger Zones}

We consider a scenario in which information about danger zones is initially unknown to the robot team, and is only revealed online as robots interact with the environment and collect clues through collaborative observation. This scenario necessitates the need for adaptive planning strategies that are dynamically adjustable as robots gather an increasing amount of knowledge of the danger sources. 
Though target tracking is the primary team-level mission, it is also necessary for robots to explore a \textit{risk map} of the environment. We assume the information gathered by any robot with proper communication will be shared with the rest of the team to improve team-level situational awareness. 

Let \( \mathcal{R}^s \) and \( \mathcal{R}^c \) denote the subset of robots with functioning sensing and communication capability, respectively. We say robots in \( \mathcal{R}^c \) are \textit{within the league} since they can communicate with each other freely. 
The set \( [p]^d \) refers to sensing danger zones known to a particular individual robot, while \( [p]^{d,c} \) denotes those shared within the league. 
Similarly, \( [q]^d \) and \( [q]^{d,c} \) represent individually known and \textit{known-within-league} communication danger zones, respectively.

If the communication league is not empty, i.e., $\mathcal{R}^c \neq \emptyset $, robots within the league are planned jointly in a centralized fashion.
Note that robots with communication capabilities may still be vulnerable to sensing attacks. In this case, the optimization problem for the league is~\cite{Liresilient2024}:
\vspace{-0.1cm}
\begin{subequations}
    \label{eq:opt_without_communication}
    \begin{align}
       & \min_{\mathbf{u}_{i,t}, \nu_i, \xi_i, \forall i \in \mathcal{R}^c } \ w_1 \sum_{i \in \mc{R}^s \cap \mathcal{R}^c} f(\mathbf{x}_{t+1}, \hat{\mathbf{y}}_{t+1})  + \\
        &\sum_{i \in\mathcal{R}^c} ( w_2 \lVert \mathbf{u}_{i,t}\rVert  + w_3 \sum_{\forall l \in [p]^{d, c}}  \| \nu_{i, l} \|  +  w_4 \sum_{\forall k \in [q]^{d,c}}  \| \xi_{i, k }\|  ) \quad \ \ \notag \\ 
        \textrm{s.t.} \
        & \mathbf{x}_{i, t+1} = \mathbf{\Phi}_i\mathbf{x}_{i, t} + \mathbf{\Lambda}_i\mathbf{u}_{i, t},  \forall  i \in\mathcal{R}^c,  \quad \\
        & \hat{\mathbf{a}}_{i, \mathcal{S}_l}^\top \mathbf{a}_{i, \mathcal{S}_l} - r_l  + \nu_{i, l}    \geq \label{eq:sensing_zone}\\
        & \quad \text{erf}^{-1}(1-2\epsilon_1)\sqrt{2\hat{\mathbf{a}}_{i, \mathcal{S}_l}^\top\mathbf{\Sigma}_{\mathcal{S}_l}\hat{\mathbf{a}}_{i, \mathcal{S}_l}} , \forall l \in [p]^{d, c}, i \in\mathcal{R}^c, \ \ \notag \\
        & \hat{\mathbf{a}}_{i, \mathcal{C}_k}^\top \mathbf{a}_{i, \mathcal{C}_k} - \delta_2 c^* +  \xi_{i, k} \geq \label{eq:comm_zone} \\
         & \quad \text{erf}^{-1}(1 - 2\epsilon_2) \sqrt{2\hat{\mathbf{a}}_{i, \mathcal{C}_k}^\top \mathbf{\Sigma}_{\mathcal{C}_k}\hat{\mathbf{a}}_{i, \mathcal{C}_k}},   \ \forall k \in [q]^{d, c}, i \in\mathcal{R}^c, \notag \\
        & \nu_{i, l}, \xi_{i, k} \in \mathbb{R}_{+}, \forall i\in\mathcal{R}^{c}, l\in [p]^{d,c}, k\in [q]^{d,c}.
    \end{align}
    \label{eq:centralized}
\end{subequations}
The definition and calculations of $\mathbf{a}_{i, \mathcal{S}_l}$, $\hat{\mathbf{a}}_{i, \mathcal{S}_l}$,
$\mathbf{a}_{i, \mathcal{C}_k}$, and $\hat{\mathbf{a}}_{i, \mathcal{C}_k}$ can be found in~\cite{Liresilient2024}. erf$(\cdot)$ is the standard error function. Eq.~\ref{eq:sensing_zone} imposes a chance-based constraint on robot $i$ to regulate its planning strategy around the sensing danger zone $\mathcal{S}_l$. Eq.~\ref{eq:comm_zone} is the chance-based constraint from the communication danger zone $\mathcal{C}_k$. We denote them in compact form for simplicity of notation as,
\begin{equation}
    \mathcal{H}(\mathbf{u}_{i, t}, \nu_{i, l}) \succeq 0, \forall i\in \mathcal{R}^{c},\forall l \in [p]^{d,c},\label{eq:chance_sensing_compact}
\end{equation}
and
\begin{equation}
    \mathcal{G}(\mathbf{u}_{i, t}, \xi_{i, l}) \succeq 0, \forall i\in \mathcal{R}^{c},\forall k \in [q]^{d,c}.
\end{equation}

The work in~\cite{Liresilient2024} discusses recovery strategies to cope with probabilistic and temporary failures and proposes a framework that dynamically switches between different optimization programs online. 
However, such approaches primarily treat failures passively and do not incorporate active estimation of danger zones. Moreover, they overlook the challenges posed by permanent failures. 
We aim to fill these gaps by developing a more generic framework that addresses both permanent and temporary failures. Our new framework achieves dynamic behavior adjustment across a wide range of failure scenarios. 

\section{Failure-aware Multi-Robot Resilient and Adaptive Target Tracking}

\subsection{Problem Formulation and Failure Modeling}
We consider two axes for classifying the types of failures: (i) temporary vs permanent, (ii) sensing and/or communication failure. This paper considers all four combinations of failure types and provides adaptive planning strategies for each case, aiming to establish a unified, resilient framework that robustly handles a diverse range of situations. We formalize the resilient multi-robot target tracking problem as the following optimization for the league with normal communication,
\begin{subequations}
    \label{eq:opt_without_communication}
    \begin{align}
       & \min_{\mathbf{u}_{i,t}, \nu_i, \xi_i, \forall i \in \mathcal{R}^c } \ w_1 \sum_{i \in \mc{R}^s \cap \mathcal{R}^c} f(\mathbf{x}_{t+1}, \hat{\mathbf{y}}_{t+1})  + \\
        &\sum_{i \in\mathcal{R}^c} ( w_2 \lVert \mathbf{u}_{i,t}\rVert  + w_3 \sum_{\forall l \in [p]^{d, c}}  \| \nu_{i, l} \|  +  w_4 \sum_{\forall k \in [q]^{d,c}}  \| \xi_{i, k }\|  ) \notag \\
        \textrm{s.t.} \
        & \mathbf{x}_{i, t+1} = \mathbf{\Phi}_i\mathbf{x}_{i, t} + \mathbf{\Lambda}_i\mathbf{u}_{i, t},  \forall  i \in\mathcal{R}^c,  \quad \\
        & \mathcal{H}(\mathbf{u}_{i,t}, \nu_{i,l}; r_{l, \text{safe}}) \succeq 0, \forall l \in [p]^{d, c}, i \in\mathcal{R}^c, \ \ \label{eq:adapt_sensing_zone} \\
        & \mathcal{G}(\mathbf{u}_{i,t}, \xi_{i,k}) \succeq 0 \ \forall k \in [q]^{d, c}, i \in\mathcal{R}^c, \label{eq:adapt_comm_zone} \\
        & \nu_{i, l}, \xi_{i, k} \in \mathbb{R}_{+},
    \end{align}
    \label{eq:adaptive_formulation}
\end{subequations}
where the objective function is made up of four terms, each weighted by $w_n,n=1, 2, 3, 4$. $w_1$ is the weight for maximizing tracking accuracy; $w_2$ is the weight for minimizing control efforts; and $w_3$ and $w_4$ are the weights for the slack variables correlated with sensing or communication danger zones, respectively. Eq.~\ref{eq:adapt_sensing_zone} is different from Eq.~\ref{eq:chance_sensing_compact} in the sense that it is now conditioned on an adaptive parameter $r_\text{safe}$, which is a desired radius for avoiding revealed sensing danger zones. $r_\text{safe}$ can be increased online when the sensing ability of the robot team is compromised. 

\subsection{Coordination Strategies Under Various Failures}

To handle both temporary and permanent failures in sensing and communication, we design a hierarchical decision-making algorithm, as specified in Algorithm~\ref{algo: framework}, that adapts the robot's behavior based on the type of failure and the communication status. Figure~\ref{fig:framework} also provides a flowchart to aid understanding of the logic. Next, we examine each branch of the algorithm and highlight key details.

\begin{algorithm}
\caption{Failure-Aware Robot Coordination}
\KwIn{a team of  $M$ robots, $attack\_type[i] \in \{temporary, permanent\}, \ \forall i \in \mathcal{R}$, $is\_connected[i] \in \{True, False\}, \ \forall i \in \mathcal{R}$}
\KwOut{Control inputs $\{\mathbf{u}[i]\}_{i \in \mathcal{R}}$}
\vspace{0.5em}
\ForEach{$i \in \mathcal{R}$}{
    \If{$attack\_type[i] = temporary$}{
        ConservativeCooldownPhase($i$)\;
        \eIf{$is\_connected[i]$}{
            $\mathbf{u}[i] \gets$ CentralizedAdaptation()\;
        }{
            $\mathbf{u}[i] \gets$ IndividualAdaptation($i$)\;
        }
    }
    \ElseIf{$attack\_type[i] = permanent$}{
        \eIf{$is\_connected[i]$}{
            EstimateCircleParameters($i$)\;
             $\mathbf{u}[i] \gets$ CentralizedAdaptation()\;
        }{
             $\mathbf{u}[i] \gets$ CircularEvasionMode($i$)\;
        }
    }
}
\label{algo: framework}
\end{algorithm}

\subsubsection{Temporary Failures}
Specifically, when a temporary failure happens, if the attacked robot remains in the communication league, its motion is dictated by the joint planning as specified in \textbf{CentralizedAdaptation} in Algorithm~\ref{algo:centralized-adaptation}; otherwise, it performs \textbf{IndividualAdaptation} when the attacked robot is isolated from the league. This essentially means that the attacked robot utilizes only its onboard resources to track targets. We assume that after the attacked robot resumes communication with its teammates, the information it has collected about danger zones will be instantly shared with the entire league. 

We set up a special \textit{cool-down} phase whenever a temporary failure occurs, regardless of whether the attacked robot still maintains communication with others. The purpose of this phase is to prevent immediate further failures and cascaded effects. This phase is immediately activated after the robot is attacked and it lasts for 10 time steps. During this period, the entire team adopts a set of extremely conservative weights, irrespective of the number of robots with normal functionality. This phase gives our system some buffer time to stabilize in the immediate aftermath of an attack.

Our centralized adaptive mechanism for the league adjusts the robots' behavior based on the number of robots that are currently under attack. When the team is mostly intact, robots can afford to take more risks to improve target tracking performance. However, as more robots become unavailable due to attacks on their sensing or communication abilities, the individual cannot contribute to the team's collective perception of targets. In response, the remaining robots gradually shift toward more conservative behaviors to mitigate further loss and to preserve longer-term autonomy. The adaptiveness is realized by changing the following in real-time based on the collective ability of robots: (i) the weights $w_1$-$w_4$ in the objective of Eq.~\ref{eq:adaptive_formulation}; (ii) $r_\text{safety}$ in Eq.~\ref{eq:adapt_sensing_zone}. 

To adjust $w_1$-$w_4$ at a particular time step $t$, we design the following rule, 
\begin{equation}
w_s = w_{s,\text{risky}} - \frac{m}{M}(w_{s, \text{risky}} - w_{s, \text{safe}}), s \in [1, 2, 3, 4],
\label{eq: safe_weight}
\end{equation}
where $M$ is the number of robots in total, $m$ is the number of robots that are attacked at time step $t$. 
Note that $w_{s,\text{risky}}$ and $w_{s,\text{safe}}$ denote the extreme values for each adaptive weight, respectively. 
This formulation ensures that when no robot is under attack, $w_s$ is essentially $w_{s,\text{risky}}$, incentivizing the system to run in the risky mode to prioritize better tracking performance. In contrast, a higher number of failures promotes prioritizing safety, sliding $w_s$ towards its other extreme value $w_{s,\text{safe}}$. 

The other aspect of the adaptive mechanism adjusts safety clearance in Eq.~\ref{eq:adapt_sensing_zone} of the sensing danger zone. Specifically, we enlarge $r_\text{safe}$ to be  $r_\text{safe}:= r_l + w_5$, where $w_5 > 0$, to further encourage robots to be conservative and stay away from the attacking source. 

We further point out that, as the isolated robot plans individually, it adopts the most conservative set of values for parameters $w_1\sim w_4$, \ie~$w_{s,\text{safe}}, s\in[1,2,3,4]$, to reduce any possible further failures. 

\subsubsection{Permanent Failures}
If instead the failure is permanent, \eg~hardware faults, we need to consider two cases separately, depending on whether it is still in normal communication with the league. If it is, we first estimate the center and radius of the danger zone, then broadcast these estimates to teammates within the league. From there, the league can perform centralized, adaptive planning. This procedure is reflected in Algorithm~\ref{algo: est_circle_center}. 

The attacked robot undergoes a \textbf{CircularMode} routine to signal its condition to the team. If it still maintains communication with the league, this robot additionally participates in \textbf{CentralizedAdaptation}, which is a scheme to adjust the weights dynamically. The rest of the team leverages this robot's circular trajectory to infer the position of the danger zone and incorporates this into centralized replanning. 

\begin{algorithm}
\caption{Centralized Adaptation}
\KwIn{Connected subgroup $\mathcal{R}^c$}
\KwOut{Control inputs $\{\mathbf{u}[i]\}_{i \in \mathcal{R}^c}$}
$M \gets$ NumberOfRobots()\;
$m \gets$ NumberOfAttackedRobots()\;
$w \gets$ ComputeSafetyWeights($M$, $m$) \texttt{// Eq.\ref{eq: safe_weight}}\; 
Solve CentralizedOptimization($\mathcal{R}^c$, $w$) \texttt{// Eq.\ref{eq:adaptive_formulation}}\; 
\label{algo:centralized-adaptation}
\end{algorithm}

\begin{algorithm}
\caption{Estimate Circle Parameters}
\KwIn{trajectory}
\KwOut{center, radius}
$center \gets$ EstimateCircleCenter(trajectory)\;
$radius \gets$ EstimateCircleRadius(trajectory)\;
Broadcast(center, radius)\;
\label{algo: est_circle_center}
\end{algorithm}

A unique challenge arises when a robot suffers a permanent communication failure. 
Since the danger zones are initially unknown and the robot can no longer broadcast information when it suffers a communication attack, the rest of the team has no direct way to learn about the threat. 
To address this, we implement a behavior-signaling mechanism: a robot that loses communication abandons its tracking task and switches to a circular motion around the attacker center, with radius equal to the threshold $\epsilon$ used in the chance constraints, shown in Algorithm~\ref{algo: circular}.
This motion pattern is designed to be easily identifiable by teammates using Algorithm~\ref {algo: est_circle_center}, allowing them to infer and avoid the hidden danger zone through observation.

\begin{algorithm}
\caption{Circular Mode}
\KwIn{robot\_id $i$}
\KwOut{Control inputs $\mathbf{u}[i]$}
$center \gets$ EstimateAttackerCenter($i$)\;
$radius \gets$ $\epsilon$\;
\While{UnderPermanentAttack}{
    ExecuteCircularTrajectory(center, radius)\;
    
}
\label{algo: circular}
\end{algorithm}

The optimization formulation itself remains consistent with prior work. The objective balances tracking performance and control effort, and soft chance constraints are used to manage risk from danger zones. The key difference is that weights $w_3$, $w_4$, and $w_5$ are now dynamically adjusted in response to the evolving status of the team. This enables the planner to adapt in real time, shifting its prioritization between performance and safety as needed.

\subsection{Target State Estimation}
In this work, we quantify the target tracking error with \textit{trace} of the estimation covariance matrix. We assume each robot is equipped with one range and one bearing sensor. Note that our adaptive replanning strategy is generic and does not hinge on the sensors used. To estimate the position of targets, we run an Extended Kalman Filter (EKF) iteratively. Next, we first introduce the classical EKF for robots within the communication league; this is followed by introducing the covariance intersection (CI) method, which merges the two streams of target estimates, with one from the communication group while the other from a previously attacked robot who temporarily lost connection and is trying to re-unite with the main group. 
\subsubsection{Extended Kalman Filtering}
The classical EKF consists of two steps: a \textit{prediction} step and an \textit{update} step. At every time $t$, in the prediction step, we compute the propagated estimate $\hat{z}_{t+1|t}$ and its corresponding covariance matrix as
 \begin{equation}
    \hat{\mathbf{y}}_{t+1|t} = \mathbf{A}\hat{\mathbf{y}}_{t}, 
    \quad 
    \mathbf{P}_{t+1|t} = \mathbf{A}\mathbf{P}_{t}\mathbf{A}^\top + \mathbf{Q},
\end{equation}
where $\mathbf{A}$ is the state transition matrix of the targets, $\mathbf{P}_t$ is the covariance matrix from the previous iteration, and $\mathbf{Q}$ is the process noise covariance matrix of targets (see Sec.~\ref{sec:background-basic}). The \textit{update step} aims at incorporating the newest measurements collected by the robots' sensor suites of the targets into its estimates, giving

\begin{align}
    \hat{\mathbf{z}}_{t+1|t+1} &= \hat{\mathbf{z}}_{t+1|t} + \mathbf{K}_{t+1}\tilde{\mathbf{z}}_{t+1|t}, \label{eq:state_update} \\
    \mathbf{P}^{-1}_{t+1|t+1} &= \mathbf{P}^{-1}_{t+1|t} + \mathbf{H}^\top \mathbf{R}^{-1}_{t} \mathbf{H}. \label{eq:cov_inv_update}
\end{align}
Here, $\mathbf{H}$ is the measurement matrix, and $\mathbf{R}$ represents the measurement noise covariance matrix of the main team. $\mathbf{K}_{t+1}$ is the Kalman gain, and $\tilde{\mathbf{z}}_{t+1|t}$ is the measurement error. Their calculations can be found in~\cite{5735231}.

\subsubsection{Covariance Intersection for Merging Estimations}
As we have discussed in previous sections, if a robot suffers from a communication attack, it loses the ability to exchange information with its teammates and thus has to perform independent target tracking (\ie single-robot target tracking), until it escapes from the danger field and resumes normal communication. By the time it attempts to rejoin the league, we have two sets of estimates of the targets' state: (i) $\hat{y}_1$ with covariance matrix $\Sigma_1$ from the main group; (ii) $\hat{y}_2$ with covariance matrix $\Sigma_2$ from the independent robot. To merge the two pieces of information together, we apply CI:
\begin{equation}
    \mathbf{\Sigma}^{-1}_\text{new} = \lambda \mathbf{\Sigma}^{-1}_{1} + (1 - \lambda) \mathbf{\Sigma}^{-1}_{2},
\end{equation}
\begin{equation}
    \hat{\mathbf{y}}_\text{new} = \mathbf{\Sigma}_\text{new} [ \lambda \mathbf{\Sigma}^{-1}_{1}\hat{\mathbf{y}}_1 + (1-\lambda)\mathbf{\Sigma}^{-1}_{2}\hat{\mathbf{y}}_{2}],
\end{equation}
to obtain a consistent new estimate. Note that the time step $t$ is left out here for simplicity. The parameter $\lambda$ can be specified using the method in~\cite{covariance_intersection}.

\begin{figure*}[ht]
    \vspace{-0.1cm} 
    \centering
    \includegraphics[width=2\columnwidth]{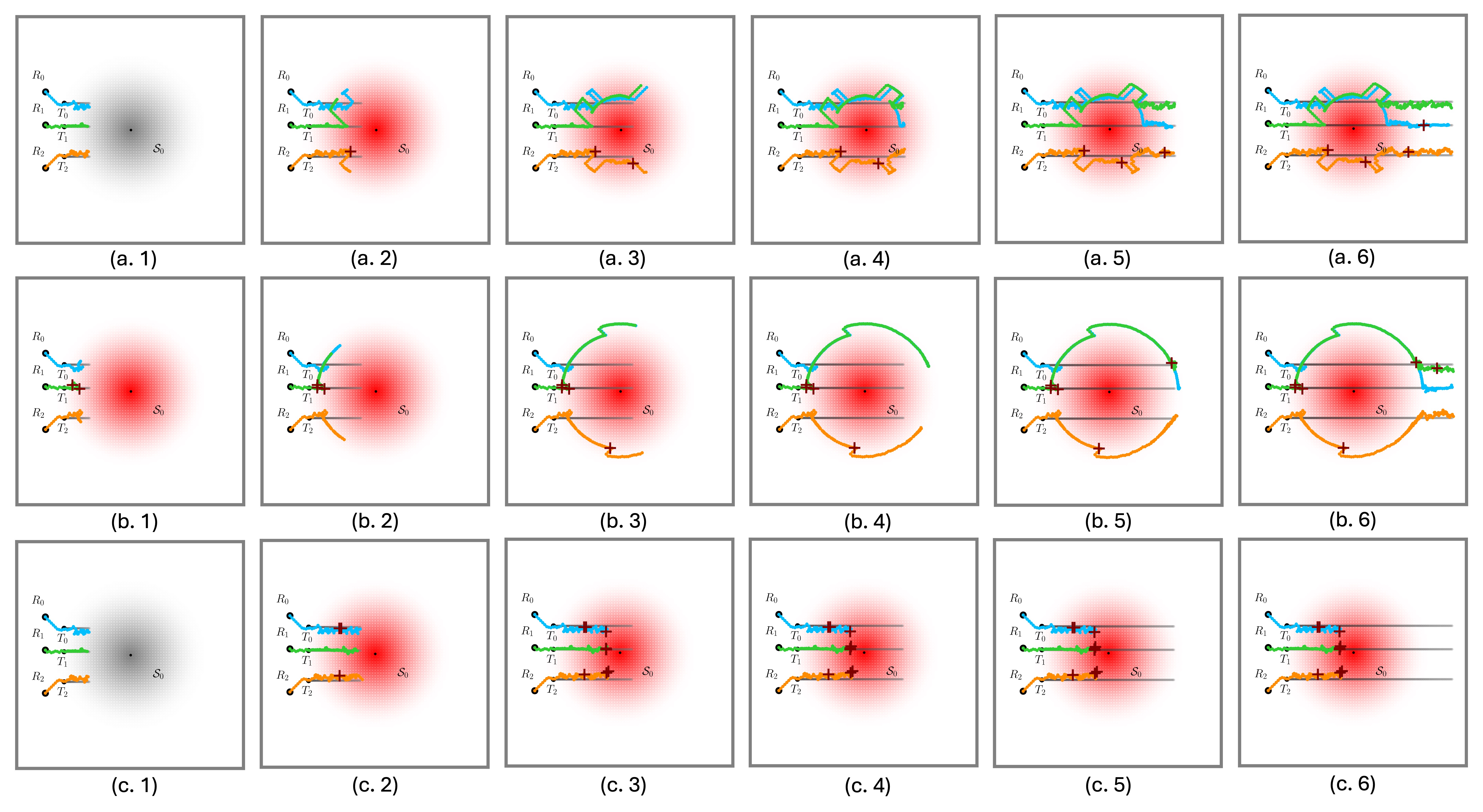}
    \vspace{-0.2cm} 
    \caption{Target tracking in the presence of a sensing danger zone. The danger zone is modeled with a mean center at [0.1, 0] and a covariance of 0.3. Subfigures (a.1)--(a.6) show adaptive tracking under temporary sensing failures; (b.1)--(b.6) depict adaptive tracking under permanent failures; and (c.1)--(c.6) illustrate the performance of the vanilla tracking approach.}
    \label{fig: sensing_attack}
\end{figure*}

\begin{figure}[htp]
\centering
\vspace{-0.00cm}
\subfigure[]{
\includegraphics[height=3.7cm]{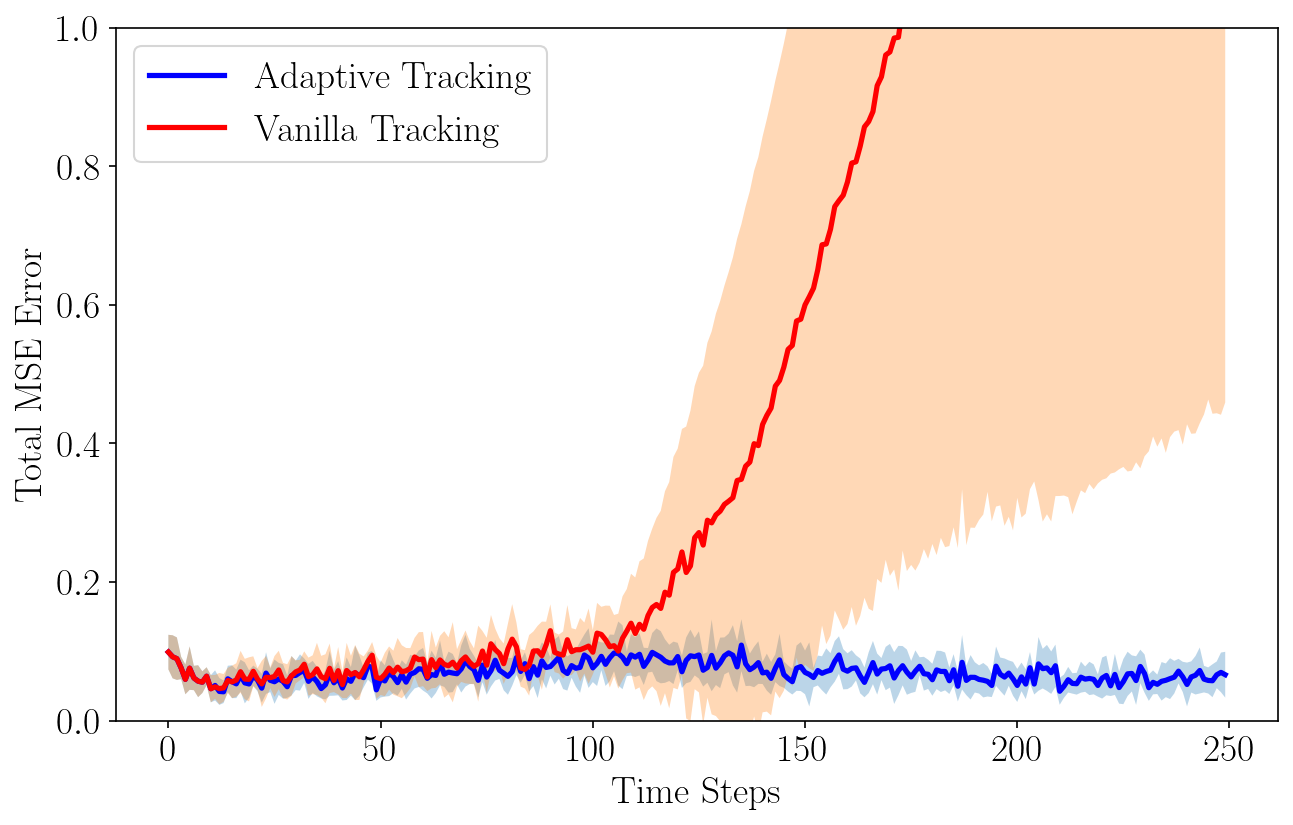}
}
\vspace{-0.38cm}

\subfigure[]{
\includegraphics[height=3.7cm]{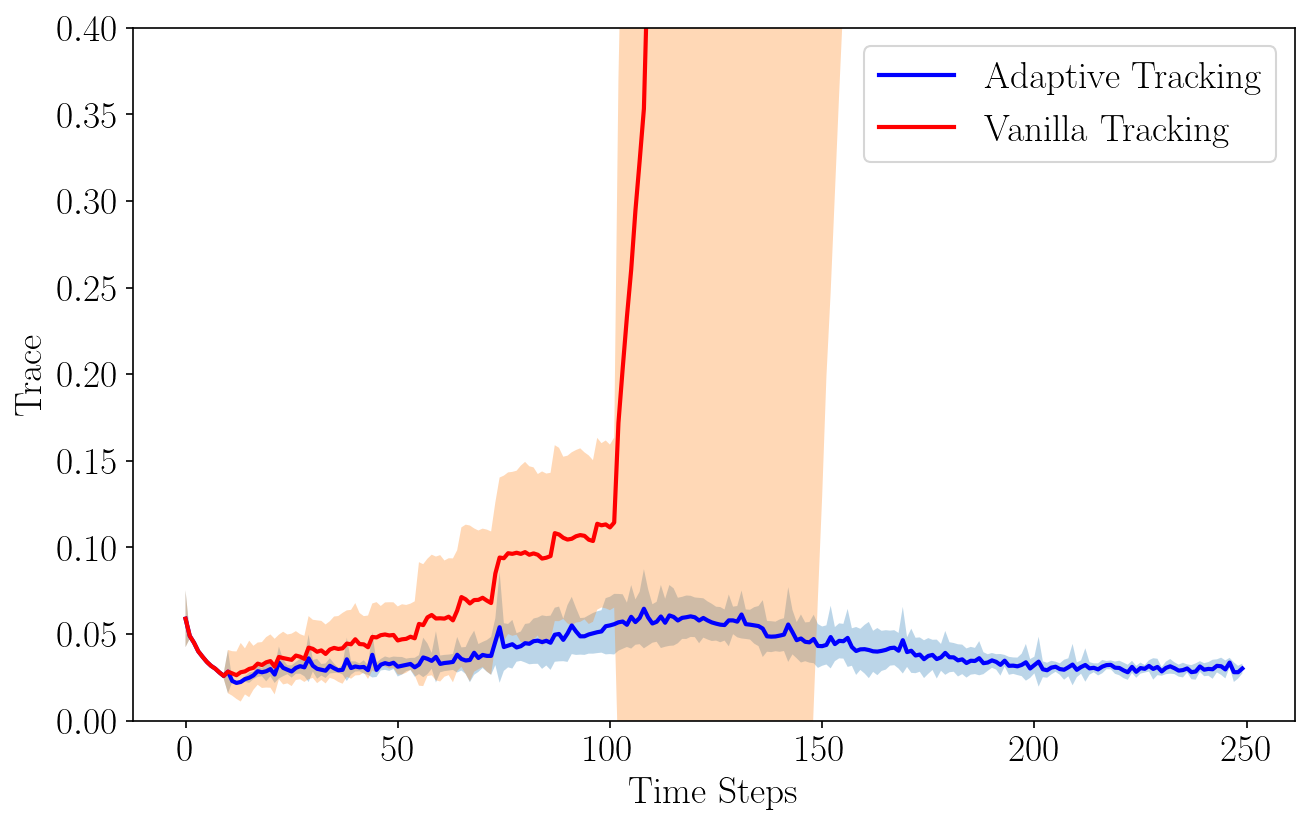}
}
\vspace{-0.2cm}
\caption{Comparison of the proposed adaptive tracking (blue) and the vanilla tracking method (red) under the scenario depicted in Figure~\ref{fig: sensing_attack}, where robots receive \textit{temporary} sensing failures. Figure (a) reports the MSE of the target state estimates, while figure (b) show the trace of the corresponding covariance matrices. Solid lines indicate the mean values, and shaded regions represent $\pm1$ standard deviation.}
\label{fig: sense_attack_temp_quant}
\vspace{-0.5cm}
\end{figure}


\begin{figure}[htp]
\centering
\vspace{-0.00cm}
\subfigure[]{
\includegraphics[height=3.7cm]{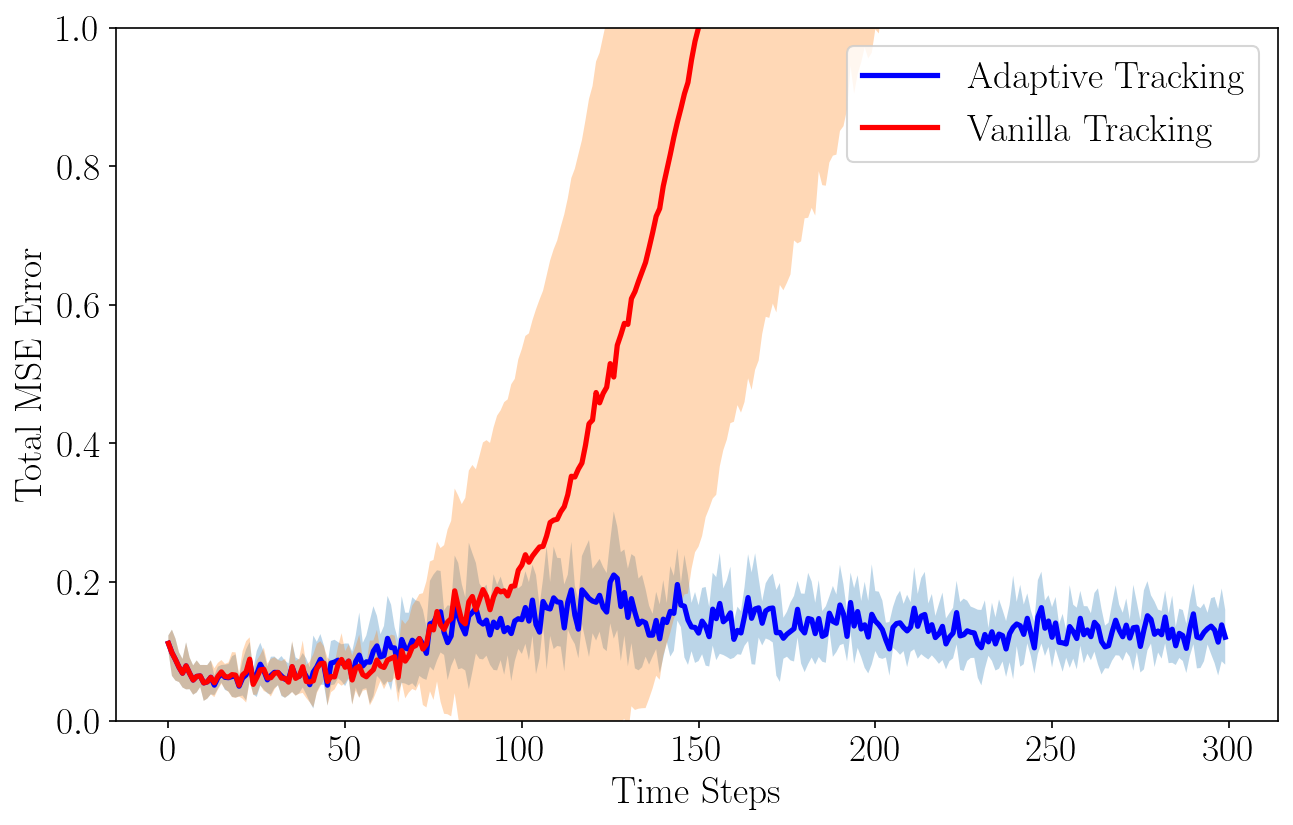}
}
\vspace{-0.38cm}

\subfigure[]{
\includegraphics[height=3.7cm]{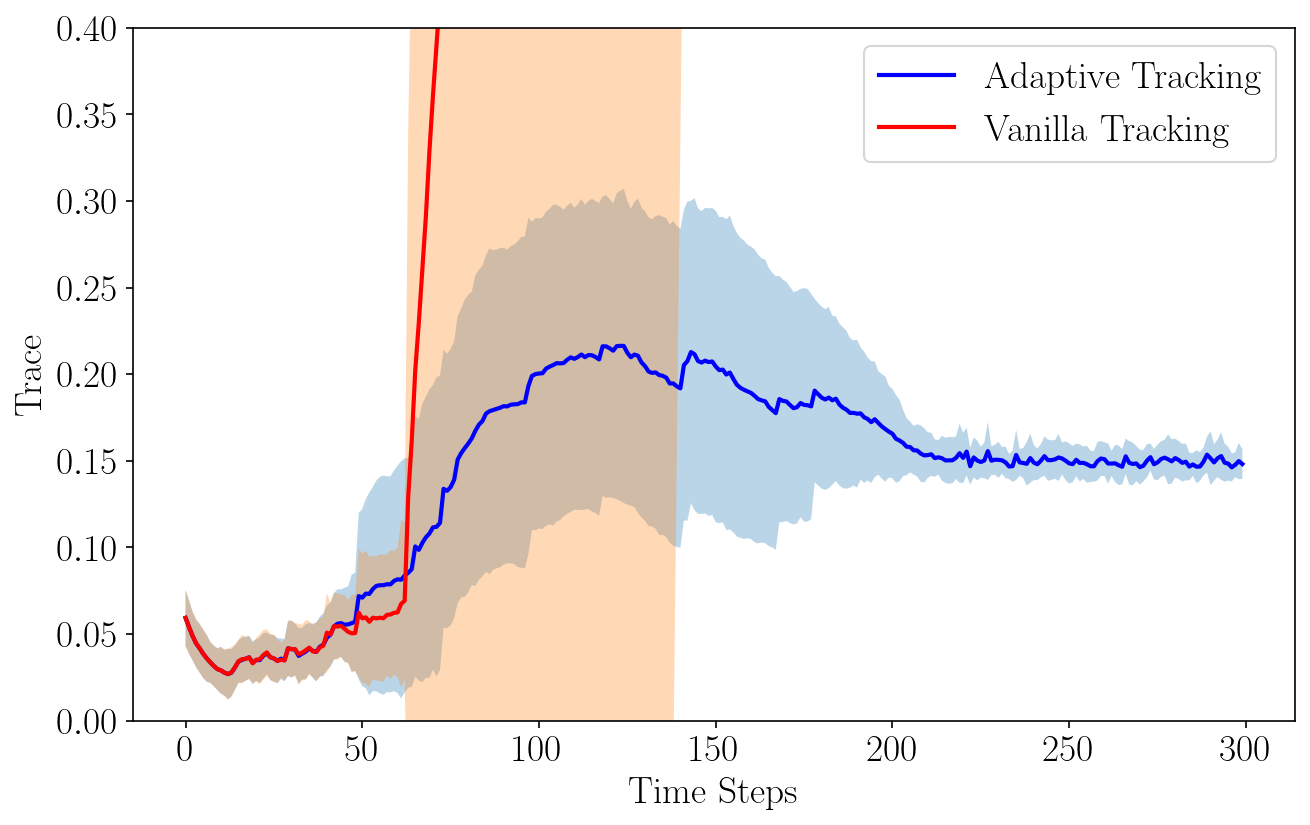}
}
\vspace{-0.2cm}
\caption{Performance of the scenarios in Figure~\ref{fig: sensing_attack} with \textit{permanent} sensing failures, showing (a) MSE, and (b) trace of covariance
matrix.}
\label{fig: sense_attack_perm_quant}
\vspace{-0.5cm}
\end{figure}

\section{Experiments}
We extensively evaluate the proposed framework through both simulation studies and real-world hardware experiments using Crazyflie drones, which are tasked with tracking multiple ground robots. 
The evaluations are designed to demonstrate the system's resilience and adaptability under various scenarios, including varying numbers of drones and ground robots, diverse settings of two distinct types of danger zones, and different failure modes. 

\subsection{Simulation}

We evaluate the adaptive tracking framework through a series of simulations that include sensing failures, communication failures, and combinations of both. Each scenario is tested under temporary and permanent disruptions to better reflect the variety of conditions robots might encounter. For comparison, we include a vanilla tracking method that does not adapt to failures. This setup enables us to assess how effectively the adaptive strategy maintains tracking accuracy when the system is subjected to various types and durations of failure. 

\subsubsection{Sensing failures}

We consider a scenario in which three robots, $\mathit{R}_0$, $\mathit{R}_1$, and $\mathit{R}_2$, are assigned to track three targets, $\mathit{T}_0$, $\mathit{T}_1$, and $\mathit{T}_2$, within an environment that contains an unknown sensing danger zone, denoted as $\mathit{S}_0$. The sensing danger zone intermittently disables a robot’s ability to observe targets, depending on its proximity to the center of the zone and the underlying failure model. We set the danger zone activation frequency to 1~Hz, with a sensing failure threshold $\delta_1 = 0.1$. Each simulation runs for 300 steps with a sampling interval of 0.1 seconds. 
Figure~\ref{fig: sensing_attack} provides the corresponding trajectories from the simulation.

\paragraph{Temporary failures}

Initially, robots have no knowledge about the sensing danger zone and focus on tracking the targets. Then $\mathit{R}_2$ encounters attack from the sensing danger zone, and immediately broadcast this information to the other team members $\mathit{R}_0$ and $\mathit{R}_1$ as shown in Figure~\ref{fig: sensing_attack}~(a.1). In response, the centralized adaptation controller adjusts the coordination weights to prioritize safety, with more conservative values computed according to the number of robots unaffected by sensing failures using Algorithm~\ref{algo:centralized-adaptation}. Once $\mathit{R}_2$ recovers, the controller reverts to a more aggressive weighting strategy to enhance tracking performance. As shown in Figure~\ref{fig: sensing_attack}~(a.2)--(a.6), the robots initially retreat from the attack zone’s center to ensure safety, and later converge again to improve tracking accuracy.

For the vanilla tracking strategy, where the robots do not react to attacks during the tracking mission as shown in (c.1)--(c.6) in Fig~\ref{fig: sensing_attack}, three robots become immobilized as they lose sensing capability due to the attack, and can no longer estimate the target state.

Figure~\ref{fig: sense_attack_temp_quant} demonstrates the mean tracking error and trace of the covariance matrix over ten tests for both the adaptive and vanilla tracking methods under temporary failures. Our adaptive tracking method maintains high tracking accuracy even in the presence of sensor attacks, whereas the vanilla tracking method fails, resulting in error and trace explosion.

\paragraph{Permanent failures}
When the sensing danger zone causes a permanent failure in a robot's sensing capability, the affected robot cannot recover from the attack. As illustrated in Figure~\ref{fig: sensing_attack}~(b.2), the centralized adaptive controller gradually adjusts the weights to become increasingly conservative as $\mathit{R}_0$ and $\mathit{R}_2$ are sequentially affected by the sensing danger zone. These adjustments help keep $\mathit{R}_1$ at a safe distance from the danger zone, preventing it from being attacked and allowing it to continue providing measurements for the team. Since permanently attacked robots remain unavailable, the set of active robots stays unchanged. As a result, the controller maintains conservative weight assignments for all robots throughout the trial, as shown in Figure~\ref{fig: sensing_attack}~(b.1)--(b.6).

Figure~\ref{fig: sense_attack_perm_quant} demonstrates the mean tracking error and trace of the covariance matrix over ten tests under permanent failures. Due to the permanent attack, the mean tracking error and trace are larger than those in the temporary failure scenario, but our proposed adaptive approach still maintains a good tracking quality compared to the vanilla method.

\subsubsection{Communication failures}
We perform similar test settings for the communication danger zones, where three robots are tracking three targets in the environment with one unknown communication danger zone, denoted as $\mathit{C}_0$. 

\begin{figure*}[!ht]
    \vspace{-0.1cm} 
    \centering
    \includegraphics[width=2\columnwidth]{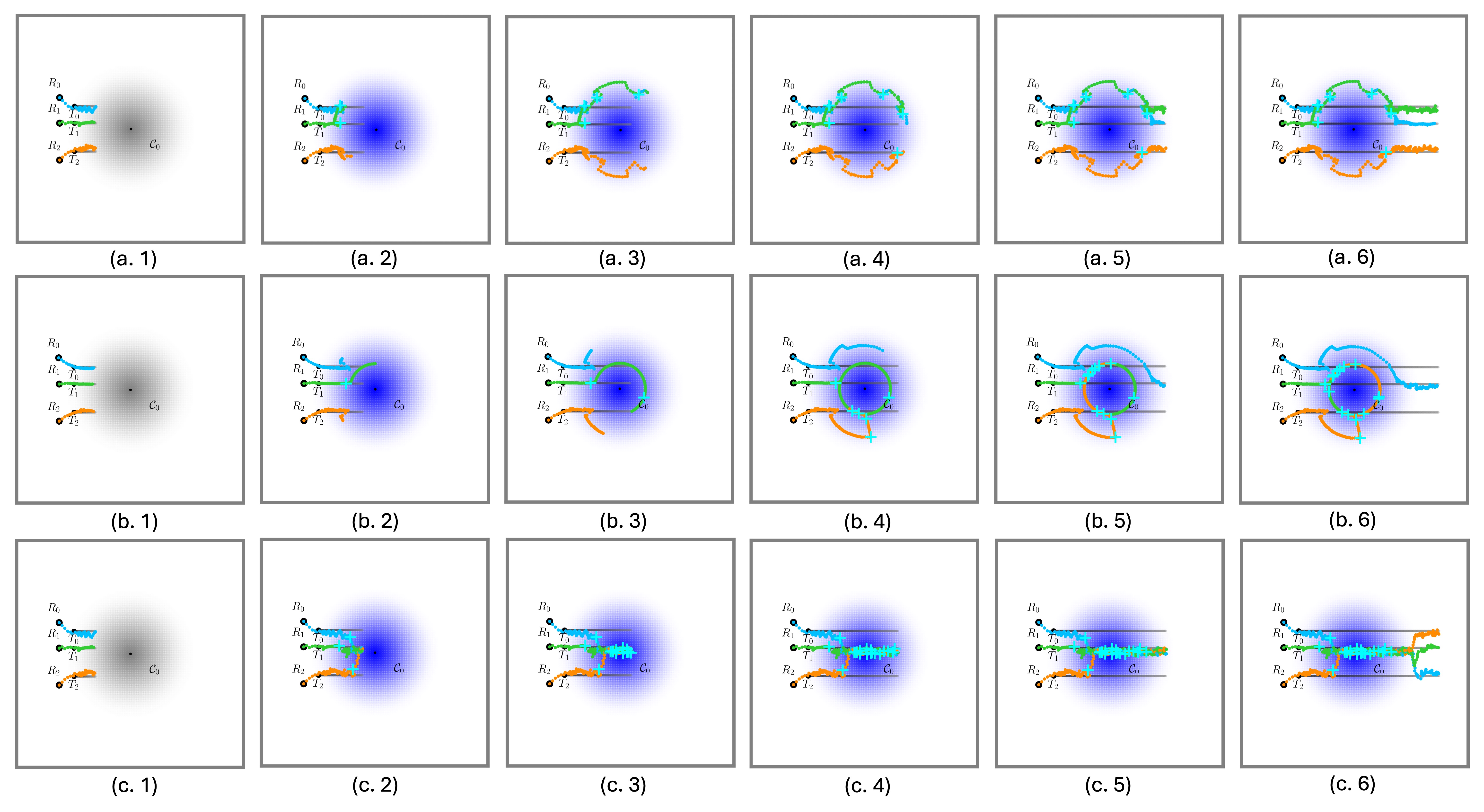}
    \vspace{-0.2cm} 
    \caption{Target tracking with a communication danger zone. (a.1)--(a.6) present adaptive strategy under temporary failures; (b.1)--(b.6) show adaptive tracking with permanent failures; and (c.1)--(c.6) show the vanilla tracking method.}
    \label{fig: comm_attack}
\end{figure*}

\paragraph{Temporary failures}
\begin{figure}[!t]
\centering
\vspace{-0.10cm}
\subfigure[]{
\includegraphics[height=4cm]{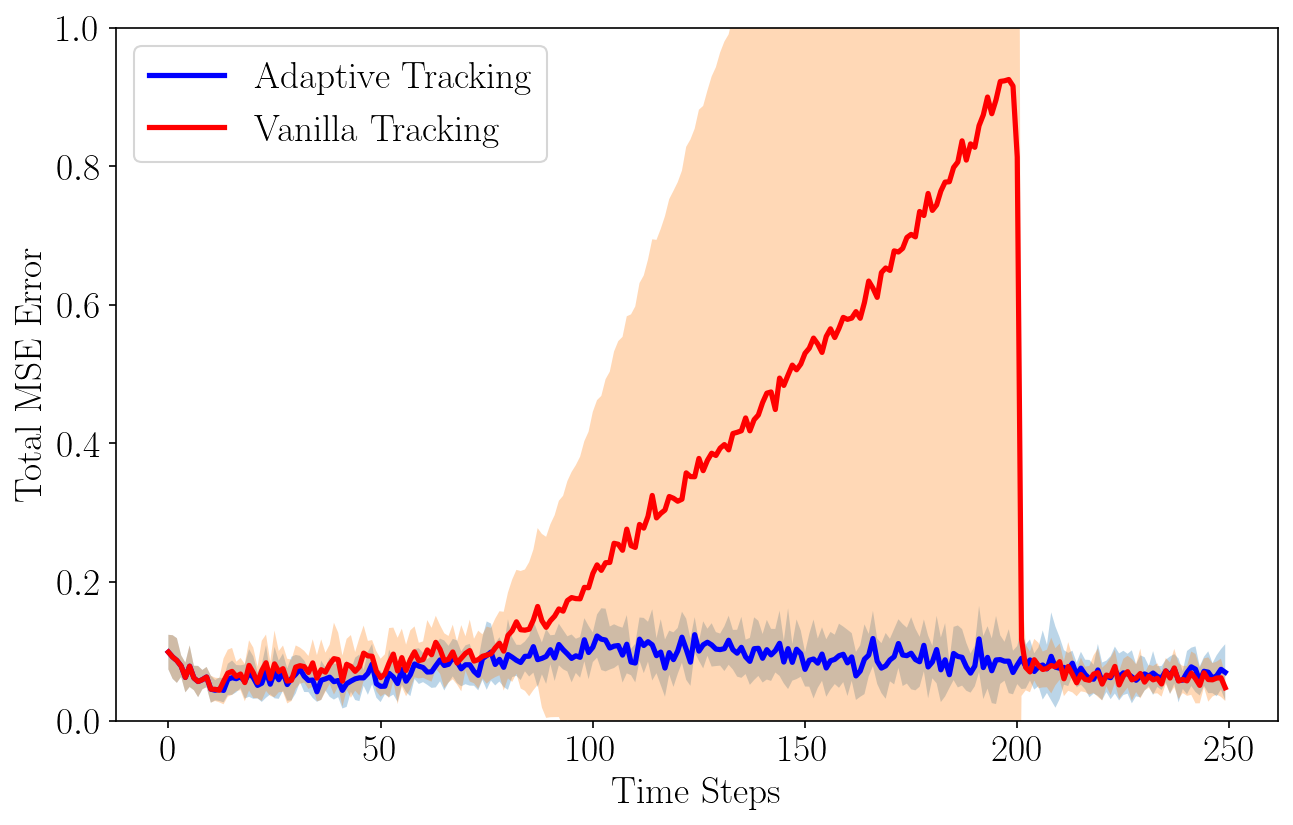}
}
\vspace{-0.3cm}

\subfigure[]{
\includegraphics[height=4cm]{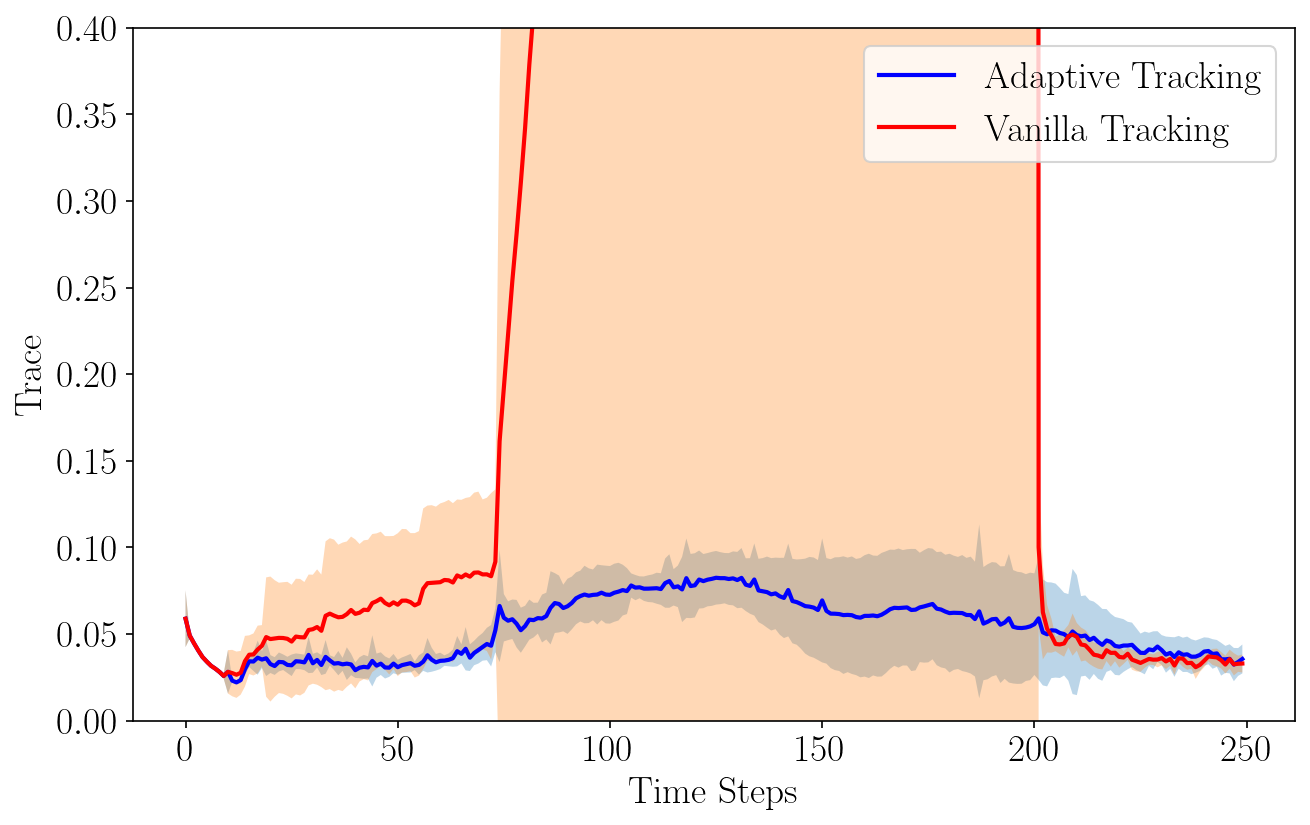}
}
\vspace{-0.2cm}
\caption{Quantitative results for experiments with \textbf{temporary} communication failures, showing (a) MSE and (b) trace of covariance matrix}
\label{fig: comm_attack_temp_quant}
\vspace{-0.2cm}
\end{figure}
As shown in (a.1)--(a.6) from Figure~\ref{fig: comm_attack}, $\mathit{R}_1$ encounters communication failures and lose connection with $\mathit{R}_0$ and $\mathit{R}_2$. $\mathit{R}_1$ first switches to an individual adaptive planner that focuses on escaping the danger zone. After $\mathit{R}_1$ recovers from the communication failure, it broadcasts the information to the rest of the team members. The rest of the team then avoids the danger zone to continue the tracking mission. The adaptive controller modifies the robots' behavior to be more conservative after receiving attacks, and then returns to a more adventurous behavior for improved tracking quality.
However, the vanilla tracking method shown in Figure~\ref{fig: comm_attack} (c.1)--(c.6) does not react to the failures and does not share information about the danger zone, resulting in robots all losing connection with each other, and collapsing towards the center to maximize the individual estimations about the target. 

Quantitative results from Figure~\ref{fig: comm_attack_temp_quant} demonstrate that the vanilla tracking method suffers from the attack and results in high estimation error and low tracking quality. It is worth noting that although all three robots recover from the communication attack using the vanilla method towards the end of the test, they are unable to provide a valuable estimate of the target during the mission. Our proposed adaptive strategy can handle failures and provide accurate estimation resiliently.

\begin{figure}[!t]
\centering
\vspace{-0.10cm}
\subfigure[]{
\includegraphics[height=4cm]{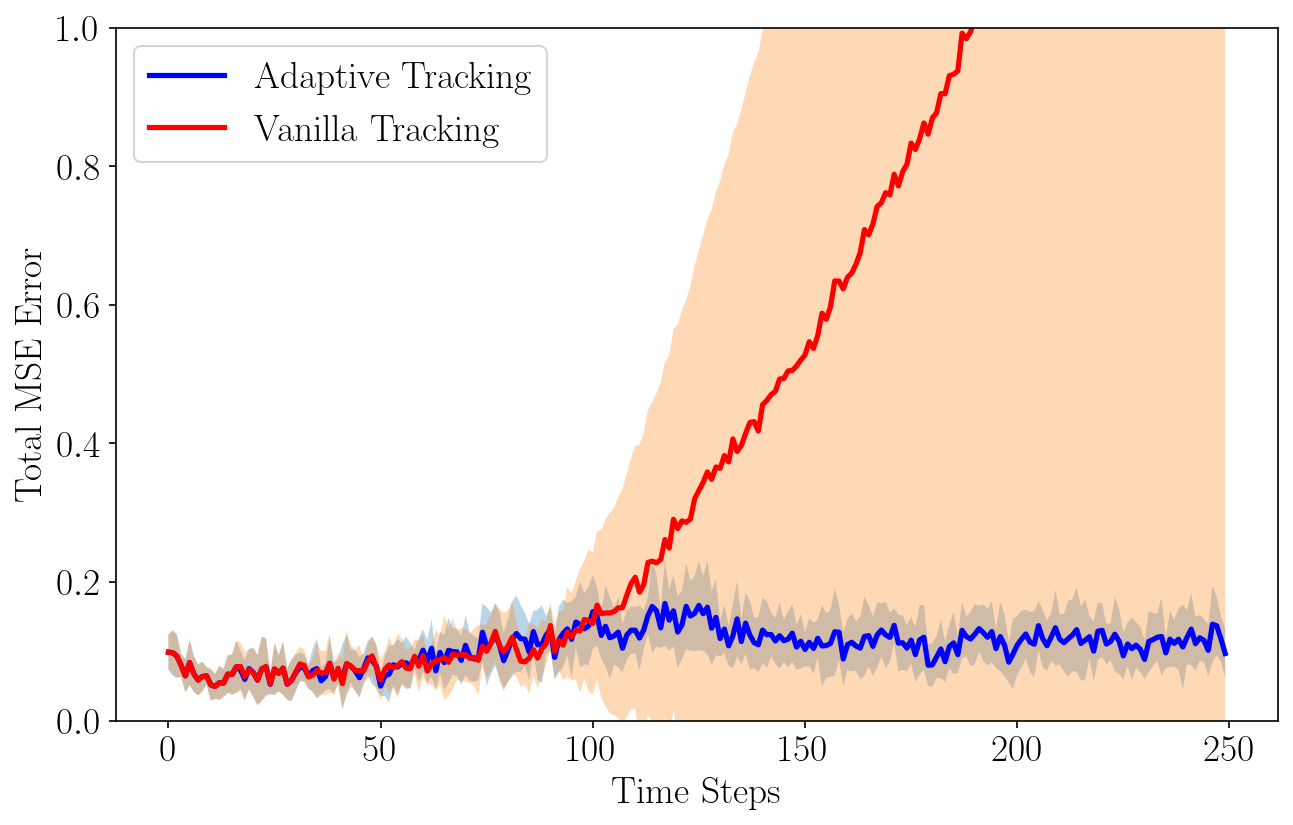}
}
\vspace{-0.3cm}

\subfigure[]{
\includegraphics[height=4cm]{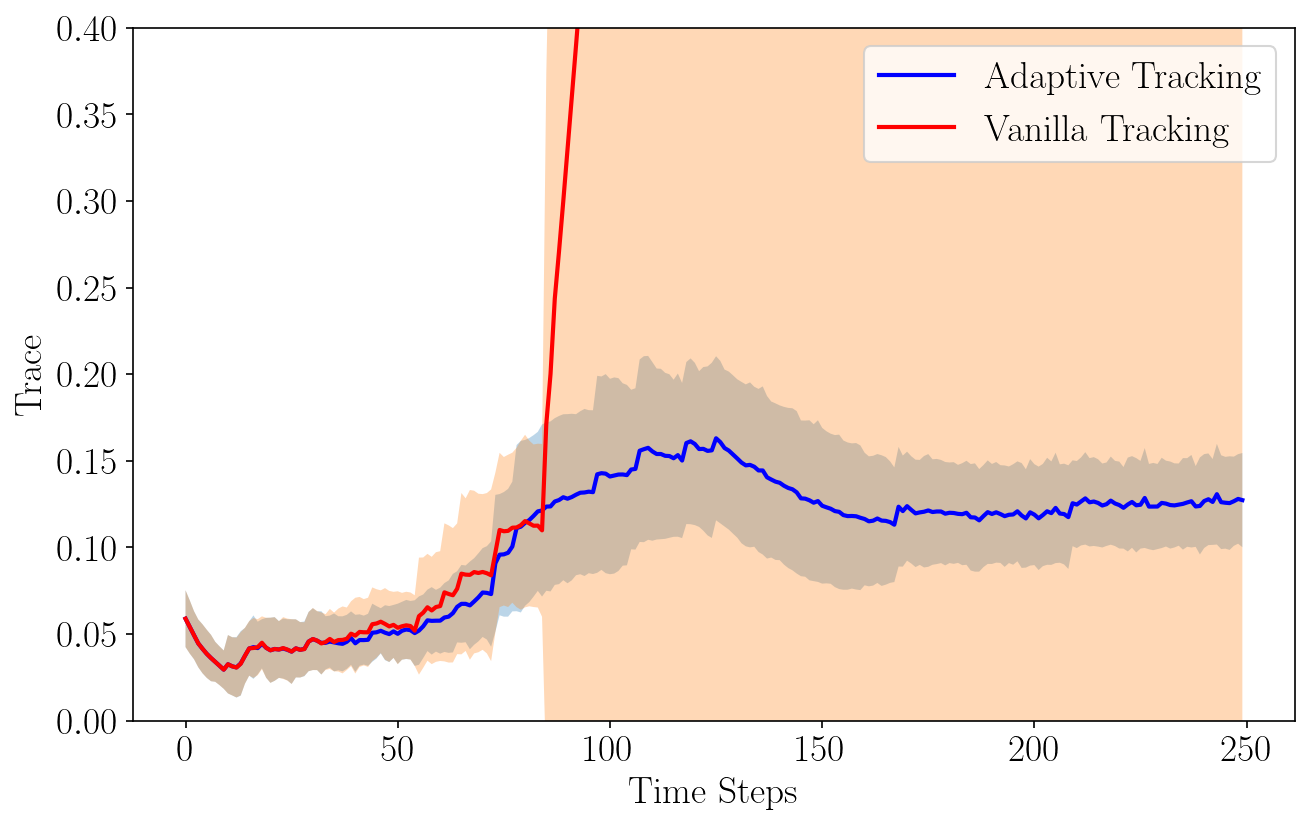}
}
\vspace{-0.2cm}
\caption{Performance under \textbf{permanent} communication failures, shown in MSE (a) and the trace of the covariance matrix (b) of the targets' state estimations.}
\label{fig: comm_attack_perm_quant}
\vspace{-0.2cm}
\end{figure}

\paragraph{Permanent failures}
When a robot becomes permanently disconnected from the communication network, it forfeits its tracking task and enters a circular motion around the communication danger zone. This behavior serves as a passive signal, enabling other team members to infer the location of the danger zone through observation. Figure~\ref{fig: comm_attack} (b.1)--(b.6) demonstrate such behavior, as $\mathit{R}_1$ starts to go around the communication danger zone when it experiences permanent communication failure. $\mathit{R}_0$ and $\mathit{R}_2$ observe the trajectory of $\mathit{R}_1$,  determine the center and region of the communication danger zone from the center and radius of the circle, and avoid the danger zone. 
Figure~\ref{fig: comm_attack_perm_quant} shows that our proposed adaptive approach works well even under permanent communication failures. Although it has a higher uncertainty in target state estimation compared to the case of temporary failures due to the irrecoverable nature of such a failure, the adaptive method still outperforms the vanilla tracking method. 

\begin{figure*}[!ht]
    \vspace{-0.1cm} 
    \centering
    \includegraphics[width=2\columnwidth]{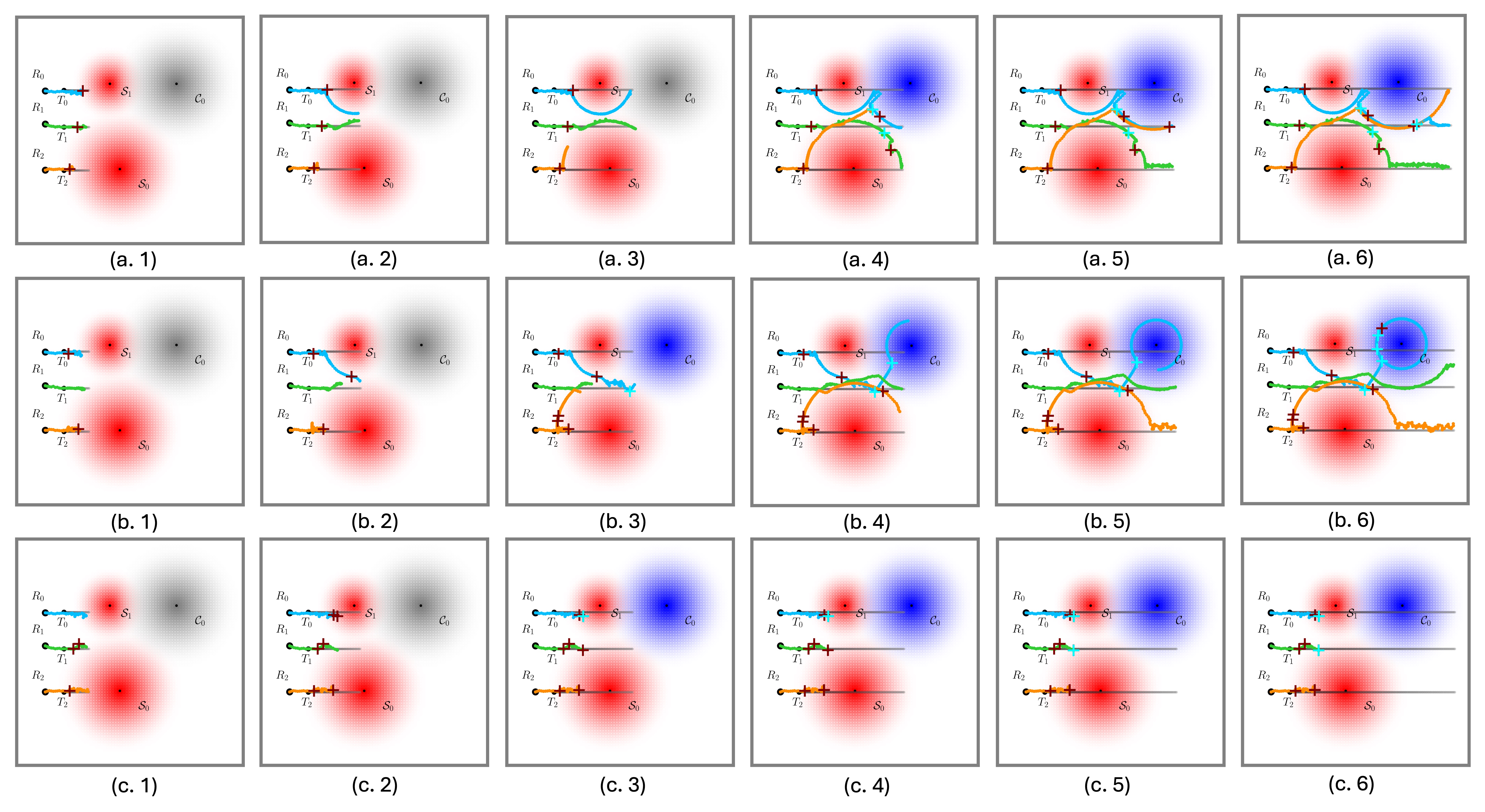}
    \vspace{-0.2cm} 
    \caption{Target tracking with combined danger zones. (a.1)--(a.6) present adaptive strategy under temporary failures; (b.1)--(b.6) show adaptive tracking with permanent failures; and (c.1)--(c.6) show the vanilla tracking method.}
    \label{fig: comb_attack}
\end{figure*}

\begin{figure}[htp]
\centering
\vspace{-0.00cm}
\subfigure[MSE]{
\includegraphics[height=4cm]{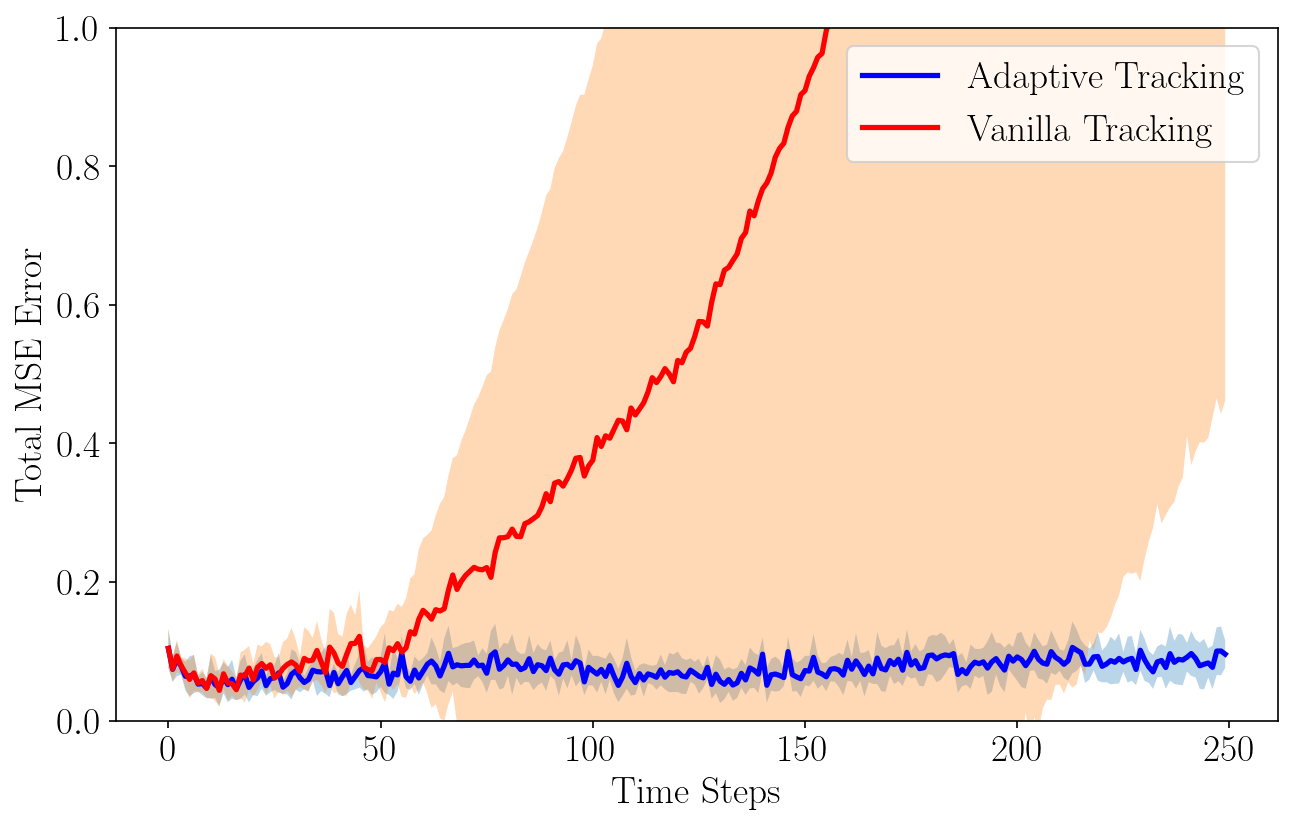}
}
\vspace{-0.3cm}

\subfigure[Trace]{
\includegraphics[height=4cm]{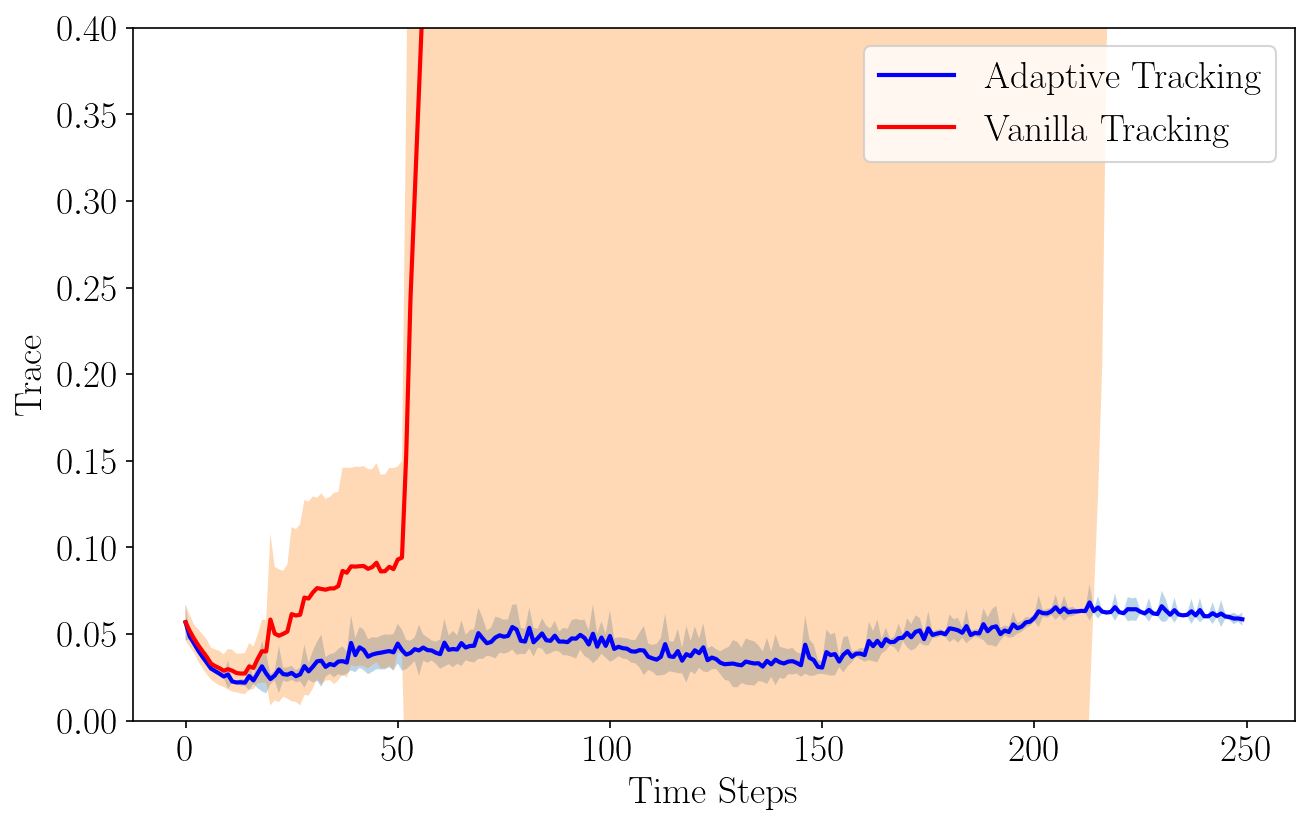}
}
\vspace{-0.2cm}
\caption{Performance of the conditions in Figure~\ref{fig: comb_attack} with \textbf{temporary} combined failures, shown in MSE (a) and the trace of the covariance matrix (b) of the targets' state estimations.}
\label{fig: comb_attack_temp_quant}
\vspace{-0.2cm}
\end{figure}

\begin{figure}[htp]
\centering
\vspace{-0.00cm}
\subfigure[MSE]{
\includegraphics[height=4cm]{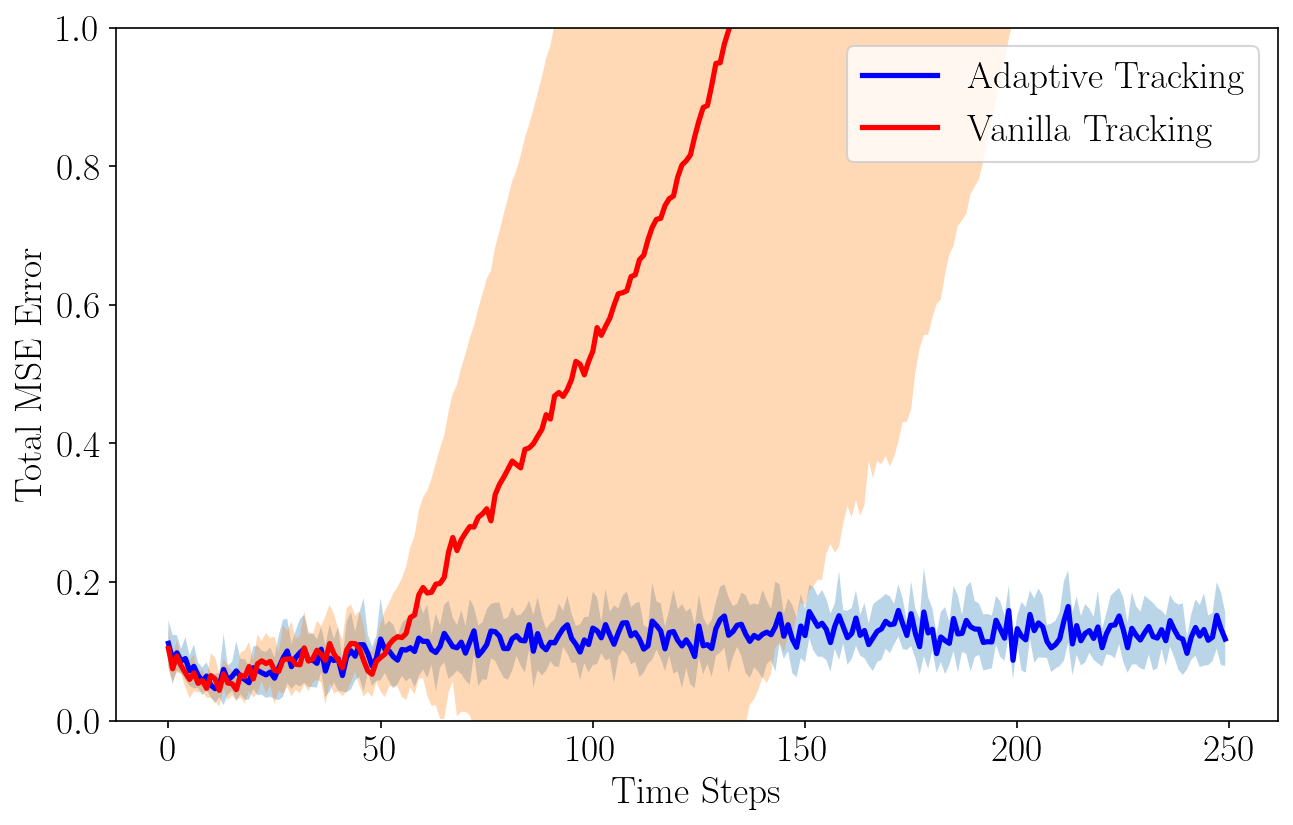}
}
\vspace{-0.3cm}

\subfigure[Trace]{
\includegraphics[height=4cm]{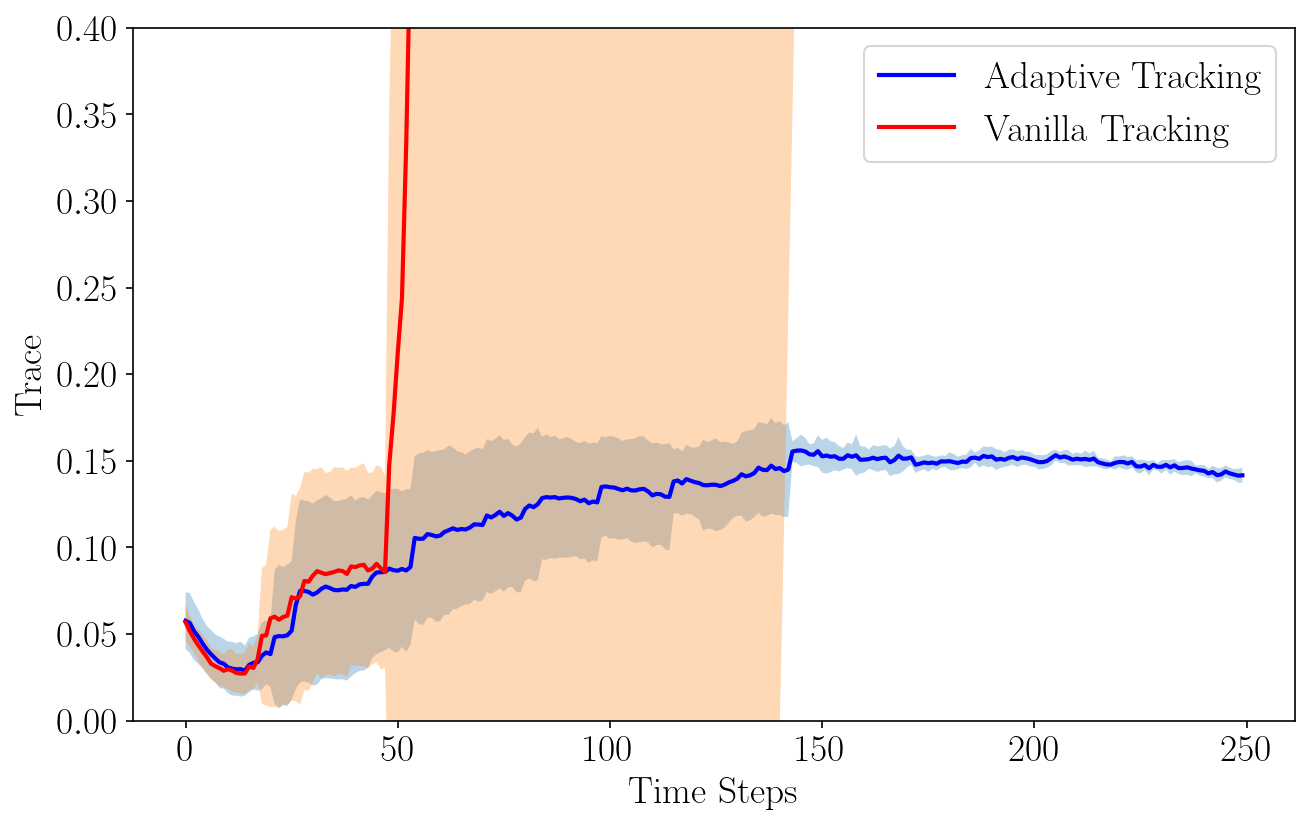}
}
\vspace{-0.2cm}
\caption{Performance under \textbf{permanent} combined failures, shown in MSE (a) and the trace of the covariance matrix (b) of the targets' state estimations.}
\label{fig: comb_attack_perm_quant}
\vspace{-0.2cm}
\end{figure}
\subsubsection{Combined failures}
We consider a more complicated scenario involving three robots tracking three targets with multiple sensing and communication danger zones in the environment. For a test, we consider two sensing danger zones with different sizes, denoted as $\mathit{S}_0$ and $\mathit{S}_1$, and one communication danger zone $\mathit{C}_0$ shown in Fig~\ref{fig: comb_attack}.

\paragraph{Temporary failures}
As shown in Figure~\ref{fig: comb_attack}~(a.1)--(a.6), the danger zones are gradually revealed to the team as individual robots encounter and recover from attacks. The adaptive controller dynamically adjusts its parameters based on the number of robots that remain unaffected, enabling the system to switch between conservative behaviors for safety and more aggressive strategies for improved tracking performance.

In contrast, the vanilla strategy depicted in Figure~\ref{fig: comb_attack}~(c.1)--(c.6) fails to maintain consistent tracking. Without the ability to share information about the danger zones or adapt to failure events, the robots are unable to avoid repeated attacks or sustain team coordination.

Quantitative results in Figure~\ref{fig: comb_attack_temp_quant} demonstrate that the proposed adaptive tracking strategy achieves superior tracking performance in terms of both accuracy and confidence. In comparison, the vanilla method is unable to complete the tracking task, as the robots become immobilized after successive attacks due to the absence of new observations.

\paragraph{Permanent failures}
As shown in Figure~\ref{fig: comb_attack}~(b.1)--(b.6), under permanent failure conditions, the robots gradually reveal the locations of the danger zones as they are attacked. Both $\mathit{R}_0$ and $\mathit{R}_2$ lose their sensing capabilities and must rely entirely on measurements provided by $\mathit{R}_1$. Using information shared by its teammates, $\mathit{R}_1$ avoids entering the danger zones and continues to support the team’s tracking efforts. Later in the experiment, $\mathit{R}_0$ suffers a communication attack and enters a circular motion pattern, passively indicating the presence of a communication danger zone. By observing this motion, $\mathit{R}_1$ estimates the center and extent of the danger zone, shares this information with $\mathit{R}_2$, and together they successfully avoid the communication hazard.

Quantitative results in Figure~\ref{fig: comb_attack_perm_quant} demonstrate that the proposed adaptive strategy is capable of resilient target tracking, even in environments with multiple sensing and communication danger zones that cause permanent failures. In contrast, the vanilla method fails to complete the tracking task, resulting in large estimation errors and significantly degraded performance when subjected to permanent failures.

\subsection{Benchmark Comparison}

\begin{figure*}[h]
    \centering
    \subfigure[]{
        \centering
        \includegraphics[width=0.63\columnwidth]{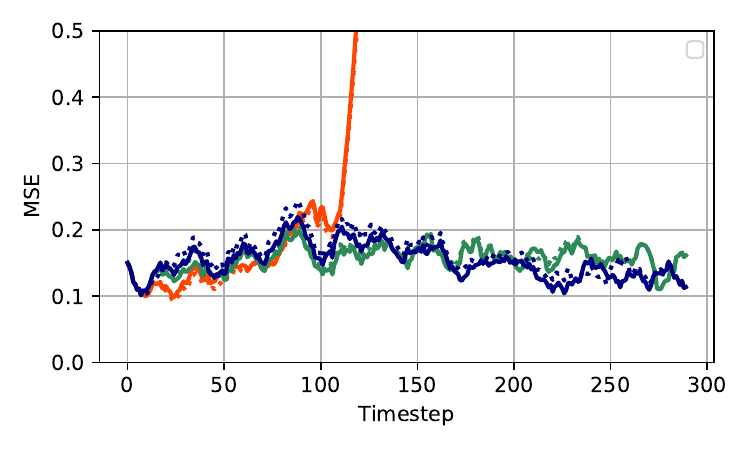}

    }
    \subfigure[]{
        \centering
        \includegraphics[width=0.63\columnwidth]{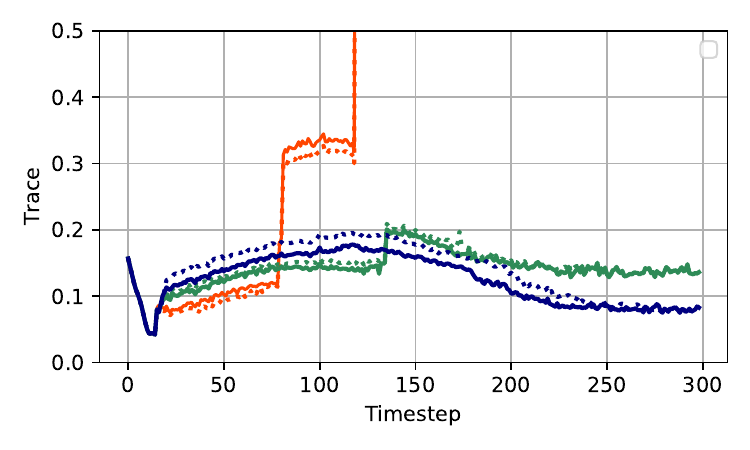}

    }
    \subfigure[]{
        \centering
        \includegraphics[width=0.63\columnwidth]{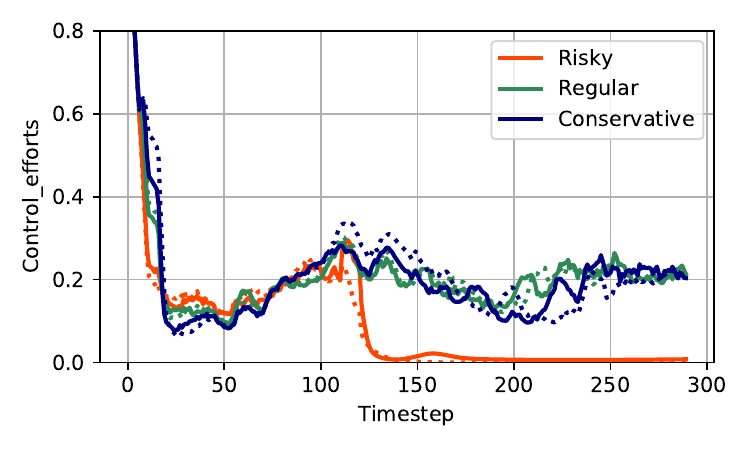}
    }
    \caption{Comparing (a) MSE, (b) trace of covariance matrix, and (c) the control efforts of our method (solid line) versus the baseline~\cite{mayya2022adaptive} (dotted line) under different scenarios.}  
    \label{fig:benchmark_quant}
\end{figure*}

The proposed framework is benchmarked against an adaptive and resilient planner in~\cite{mayya2022adaptive}. Note that~\cite{mayya2022adaptive} defines a safety-aware observability Gramian (SOG) to encode the abundance and diversity of sensors available, and the planning algorithm hinges on this SOG value. For the sake of fairness of comparison, only permanent sensor failures are considered for benchmarking, as this is the only scenario considered in~\cite{mayya2022adaptive}. The benchmark experiments consist of three parts. The first part shows that our approach is equally capable as~\cite{mayya2022adaptive} in terms of catering for a diverse set of safety requirements. The second part examines how the number of robots affects the target tracking performance of both our approach and the baseline. In the third part, we increase the complexity of the environment by adding more sensing danger zones and investigate how longer-term autonomy is achieved. Next, we show the details and results for each part.

\subsubsection{Diverse Adaptive Behaviors}
The acceptable risk level in our approach is controlled by $\epsilon_1$, while in~\cite{mayya2022adaptive}, it is adjusted by parameter $\rho_2$ (see Eq.~17e there). We aim to demonstrate that by adjusting the hyperparameters $\epsilon_1$ and $\rho_2$, our approach can exhibit an equally diverse range of adaptive behaviors as~\cite{mayya2022adaptive}, and thus is widely applicable regardless of whether end-users are lenient or stringent with safety requirements. 

To do this, we adopt three settings, which correspond to the robots' behaviors being \textit{risky, regular, or conservative}. Risky means that robots have a higher acceptable risk level, while conservative means the vice versa. The parameter values we used are specified in Table~\ref{tab:benchmar_parameters}. These parameters are hand-tuned. The environment initially contains a single unknown sensing danger zone, and four robots are tasked with tracking four targets. The results are shown in Figure~\ref{fig:benchmark_quant}. We compare the MSE, trace of the covariance matrix, as well as the norm of the control input throughout the tracking procedure. We use different colors to showcase different levels of acceptable risk. All solid lines represent our proposed approach, while dotted lines are for the method in~\cite{mayya2022adaptive}. The curves are smoothed for visualization purposes. 

Note that the goal here is not to compare the tracking error itself, but to demonstrate the ability of our approach to exhibit diverse behaviors under different safety requirements. We observe that under the risky setting, both approaches experience serial sensor failures quickly, which causes the tracking error to increase rapidly after approximately 120 steps. Without the sensing ability, the robot team can no longer track the targets, so they essentially remain stationary, which explains the nearly zero control efforts. In contrast, in either the regular or conservative case, the robots sustain a finite estimating error as they follow the targets. 

\begin{table}[h]
\centering
\caption{Parameter Values}
\label{tab:benchmar_parameters}
\begin{tabular}{lccc}
\toprule
\textbf{Method} & \textbf{$\epsilon_1$ in ours} & \textbf{$\rho_2$ in~\cite{mayya2022adaptive}}  \\
\midrule
Scenario 1 (Risky)   & 0.05   & 0.6  \\
Scenario 2 (Regular)  & 0.02   & 0.3            \\
Scenario 3 (Conservative)  & 0.01   & 0.2  \\
\bottomrule
\end{tabular}
\end{table}

\subsubsection{Tracking with a varying number of robots}

\begin{figure}
    \centering
    \centering
    \includegraphics[width=\columnwidth]{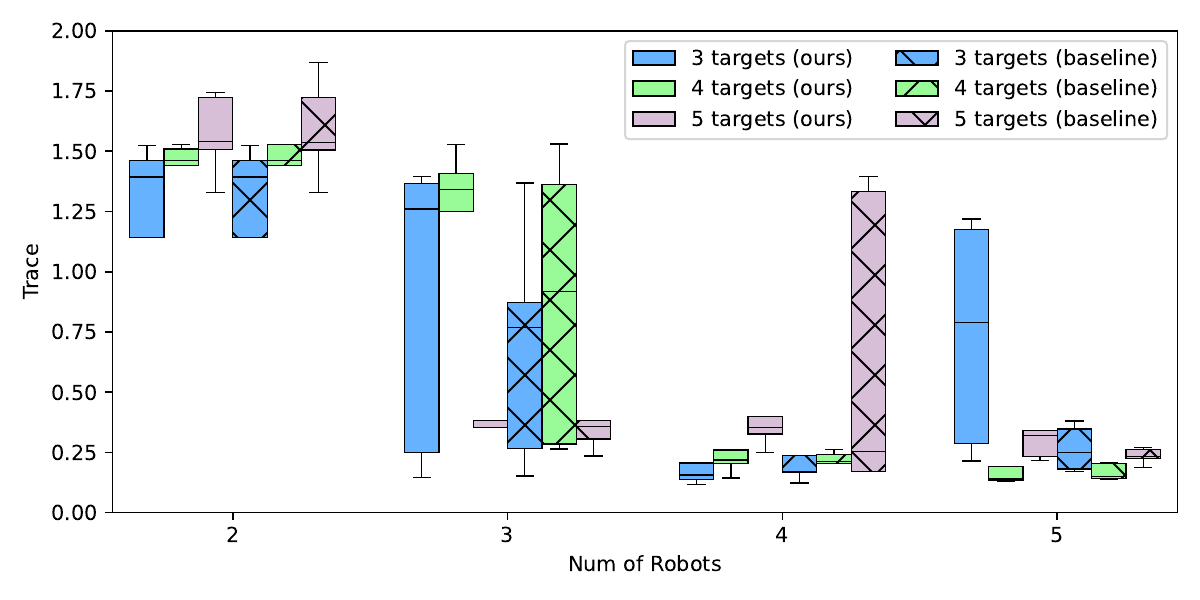}
    \caption{
    Comparison of the trace with varying numbers of robots (2 to 5) and targets (3, 4, and 5). 
    Results are shown for both the proposed approach and the baseline method in~\cite{mayya2022adaptive} (with hatched patterns) across different target scenarios. 
    Lower trace values indicate better performance. 
    }
    \label{fig:benchmark_vary_robots_and_targets}
\end{figure}

We further study whether our approach exhibits similar behavior as~\cite{mayya2022adaptive} as we change both the number of robots and the number of targets. Specifically, in an environment with two sensing danger zones, we vary the number of robots from 2 to 5 to track 3-5 targets, and report the overall trace error. For each combination of robots and targets, we repeat the simulation 5 times (each with a different random seed). We record the trace averaged across the temporal dimension for each trial and report the aggregate results in the form of box plots in Figure~\ref{fig:benchmark_vary_robots_and_targets}. As expected, as we increase the number of robots, the overall tracking quality improves, as evidenced by the lower trace error, since the robot team is more abundantly resourced. At the same time, increasing the number of targets generally increases the tracking error. Note that although we have repeated 5 times for each combination of the number of robots and targets, there still exists a non-negligible variance in the overall trace values for some cases (\eg 3 robots tracking 3 targets), resulting in non-monotone patterns that we see.

\begin{figure}
    \centering
    \includegraphics[width=0.98\linewidth]{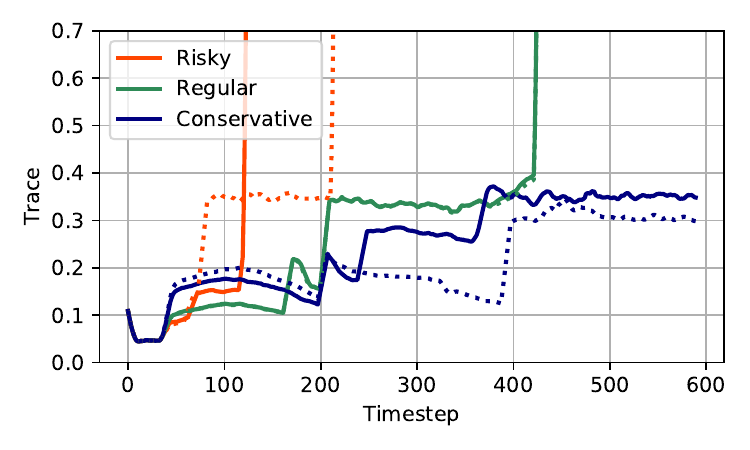}
    \caption{Benchmarking in an environment with 3 unknown sensing danger zones. Solid lines correspond to our method while dotted lines are for the baseline in~\cite{mayya2022adaptive}. }
    \label{fig:benchmark_complex_env}
\end{figure}
\begin{figure*}[!htbp]
    \vspace{-0.1cm} 
    \centering
    \includegraphics[width=2\columnwidth]{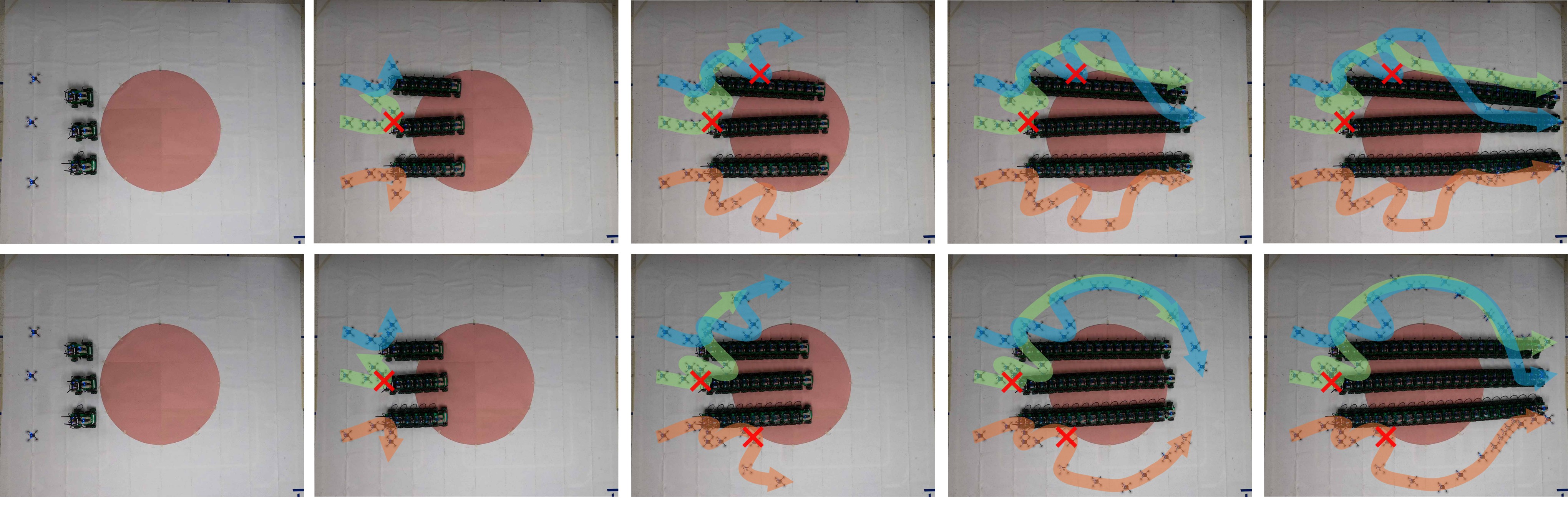}
    \caption{Snapshots at different time steps showing three robots tracking three targets in the presence of one unknown sensing danger zone. The top row illustrates scenarios with temporary sensor attacks, while the bottom row depicts cases of permanent sensor failures. Sensor attacks are marked with {\color{red}{\ding{53}}}.
    }
    \label{fig: sense-real}
\end{figure*}

\begin{figure*}[!ht]
    \vspace{-0.1cm} 
    \centering
    \includegraphics[width=2\columnwidth]{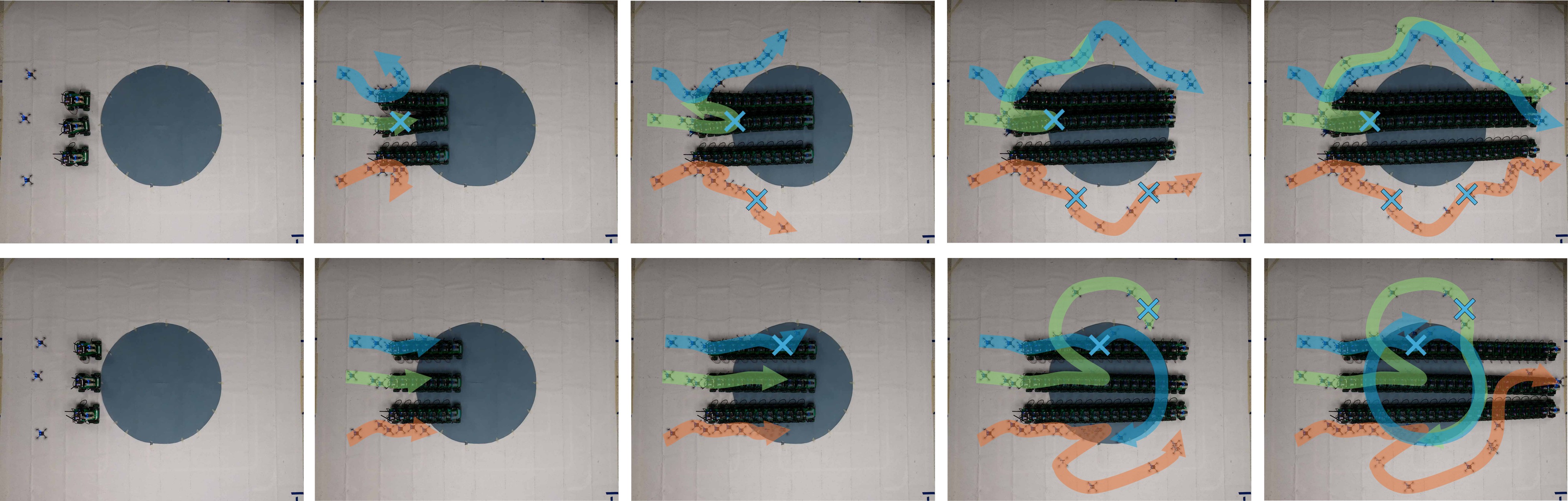}
    \caption{Snapshots at different time steps showing three robots tracking three targets with one unknown communication danger zone under temporary (top) and permanent (bottom) failures. Communication attacks are marked by {\color{cyan}{\ding{53}}}.}
    \label{fig: comm-real}
\end{figure*}
\subsubsection{Operation in a complex environment}

Finally, we run both approaches in a more complex environment with 3 sensing danger zones under different levels of risk requirements. We simulate the motion of the targets and robots for up to 600 steps. The trace error is reported in Figure~\ref{fig:benchmark_complex_env}. After around 100 steps, the robot team encounters the first sensing danger zone. If we set a higher acceptable risk level, meaning that the robots are more willing to take the risk, they experience severe failures and thus can no longer finish the task. The case with a medium safety requirement persists until after 400 time steps, while the conservative case is the only one in which robots successfully follow the targets and survive through all three danger zones. This demonstrates that in an environment with multiple sensing danger zones, a stringent safety requirement is necessary for both our approach and the method in~\cite{mayya2022adaptive} to achieve long-term autonomy. 

\subsection{Real-World Tests}

To further demonstrate the resiliency and adaptiveness of our proposed framework, we conduct a series of hardware experiments in diverse environments, featuring initially unknown hazards. For hardware evaluation, we use Crazyflie drones~\cite{crazyflie} as tracking robots and Yahboom ROSMASTER X3 ground vehicles~\cite{yahboom} as targets. Figure~\ref{fig: sense-real} presents the trajectories of three robots tracking three targets in an environment containing a single, initially unknown sensing danger zone. The top and bottom rows correspond to scenarios involving temporary and permanent sensing attacks, respectively. In the case of temporary attacks, the proposed controller adaptively adjusts the coordination weights based on the number of robots that remain unaffected by the danger zone, effectively balancing tracking accuracy and safety. As shown in the trajectories, the robots initially move away from the danger zone upon detecting new attacks to ensure safety. Once the affected robot recovers, the team reapproaches the targets to improve tracking performance. In contrast, when robots suffer permanent attacks, they are unable to recover, prompting the team to continuously apply safety-prioritized coordination weights. This leads to more conservative trajectories, with the robots maintaining a safe distance from the danger zone throughout the task. 

Figure~\ref{fig: comm-real} and \ref{fig:real_exp} show the trajectories of robots tracking targets under communication and combined danger zones, respectively.  In both figures, the top rows correspond to scenarios involving temporary attacks, while the bottom rows illustrate cases of permanent attacks. Similar to the results in Figure~\ref{fig: sense-real}, the proposed adaptive controller dynamically balances safety and tracking quality based on the subset of robots that remain functional. 
Notably, when a robot is permanently affected by a communication danger zone, it loses the ability to inform teammates of the attacker's location due to its inability to recover and communicate. In such cases, the robot abandons the tracking task and instead performs a circular motion around the estimated center of the danger zone. This behavior serves as a passive signal, allowing the remaining robots to observe the trajectory and infer the location of the communication hazard.

\section{Conclusion}
In summary, this work proposes a novel failure-aware adaptive target tracking framework that enables multi-robot teams to operate resiliently in environments with unknown sensing and communication danger zones. By dynamically adjusting coordination strategies based on the availability and status of team members, the proposed approach achieves a balance between safety and tracking performance under both temporary and permanent failures. Extensive simulations and hardware experiments validate the framework's effectiveness across a range of scenarios.

Future work will explore tracking in more adversarial settings, where targets may actively evade robots or even cooperate with danger zones to disrupt the mission. 
We also plan to investigate scenarios involving mobile or dynamic danger zones, which introduce additional challenges for situational awareness and adaptation. 
Furthermore, incorporating more complex and realistic failure models, such as partial degradation, stochastic fault propagation, and multimodal attack patterns, will enhance the robustness and applicability of the system in complex, real-world environments. 
These extensions will further advance the development of resilient and intelligent multi-robot systems capable of operating autonomously in uncertain and hostile environments.

\bibliography{references}

\end{document}